%% file: main.tex
\pdfoutput=1

\documentclass[11pt]{article}

\usepackage{arabtex}
\usepackage{utf8}

\usepackage{times}
\usepackage{latexsym}
\usepackage{booktabs}
\usepackage{xargs}
\usepackage{tikz-dependency}
\usepackage[utf8]{inputenc}
\usepackage[main=english, arabic, russian]{babel}
\usepackage{CJKutf8}

\usepackage{CJKutf8}
\usepackage{todonotes} 
\usepackage{caption}
\usepackage{subcaption}
\usepackage{changes}
\usepackage{paralist}

\usepackage{tablefootnote}

\usepackage{naacl2021}

\usepackage{times}
\usepackage{latexsym}

\usepackage[utf8]{inputenc}

\usepackage{microtype}

\newcommand{\name}[0]{\textsc{xSID}}

\title{From Masked Language Modeling to Translation: Non-English Auxiliary
Tasks Improve Zero-shot Spoken Language Understanding}

\author{
\begin{minipage}{\textwidth}\centering
Rob van der Goot$^1$, 
Ibrahim Sharaf$^2$, 
Aizhan Imankulova$^3$, 
Ahmet {\"U}st{\"u}n$^4$, \\
Marija Stepanovi{\'c}$^1$, 
Alan Ramponi$^{5,6}$, 
Siti Oryza Khairunnisa$^3$, 
Mamoru Komachi$^3$, 
Barbara Plank$^1$\\
\end{minipage}
\vspace{.1cm}
\\
\begin{minipage}{\textwidth}\centering 
$^1$IT University of Copenhagen,
$^2$Factmata,
$^3$Tokyo Metropolitan University,
$^4$University of Groningen,
$^5$University of Trento,
$^6$Fondazione The Microsoft Research -- University of Trento Centre for Computational and Systems Biology (COSBI)
\end{minipage}
\vspace{.05cm}
\\
\begin{minipage}{\textwidth}\centering 
\texttt{\{robv,bapl\}@itu.dk}
\end{minipage}
}

\date{}

\begin{document}
\maketitle

\begin{abstract}
The lack of publicly available evaluation data for low-resource languages
limits progress in Spoken Language Understanding (SLU). As key tasks like
intent classification and slot filling require abundant training data, it is
desirable to reuse existing data in high-resource languages to develop models
for low-resource scenarios.  We introduce \name{}, a new benchmark for
cross-lingual (\textsc{x}) Slot and Intent Detection in 13 languages from 6
language families, including a very low-resource dialect. To tackle the
challenge, we propose a joint learning approach, with English SLU training data
and non-English auxiliary tasks from raw text, syntax and translation for
transfer. We study two setups which differ by type and language coverage of the
pre-trained embeddings. Our results show that jointly learning the main tasks
with masked language modeling is effective for slots, while machine translation
transfer works best for intent classification.\footnote{The source code,
dataset and predictions are available at:
\url{https://bitbucket.org/robvanderg/xsid}}
\end{abstract}

\section{Introduction}
\input{intro}

\section{Data}\label{sec:sec2-data}
\input{data}

\section{Models}\label{sec:model}
\input{models}

\section{Results}\label{sec:results}
\input{results}

\section{Analysis}\label{sec:analysis}
\input{analysis}

\section{Related Work}
\input{relwork}

\section{Conclusions}\label{sec:conclusion}
\input{conclusion}

\section*{Acknowledgements}
We would like to thank Yiping Duan, Kristian N\o{}rgaard Jensen, Illona
Flecchi, Mike Zhang, and Caroline van der Goot for their annotation efforts. We
thank Dennis Ulmer for his help with significance testing. Furthermore, we
thank Fabian Triefenbach, Judith Gaspers and the anonymous reviewers for the
feedback. We also thank NVIDIA, Google cloud computing and the ITU
High-performance Computing cluster for computing resources. This research is
supported in part by the Independent Research Fund Denmark (DFF) grant
9131-00019B and 9063-00077B and an Amazon Faculty Research (ARA) Award.

\bibliography{NLU}
\bibliographystyle{acl_natbib}

\clearpage
\appendix
\input{appendix}

\end{document}

%% file: intro.tex
Digital conversational assistants have become an integral part of everyday life
and they are available, e.g., as standalone smart home devices or in
smartphones. Key steps in such task-oriented conversational systems are
recognizing the \textit{intent} of a user's utterance, and detecting the main
arguments, also called \textit{slots}. For example, for an utterance like ``Add
reminder to swim at 11am tomorrow'', these key Natural Language Understanding
(NLU), or Spoken Language Understanding (SLU) tasks are illustrated in
Figure~\ref{fig:sentence}.  As slots depend on the intent type, leading models
typically adopt joint solutions~\cite{chen_bert_2019,qin_multi-domain_2020}.  

\definecolor{movie}{HTML}{FF7F00}
\definecolor{object}{HTML}{27ae60}

\begin{figure}
\centering
\begin{tabular}{|l|}
\hline
Add reminder to \colorbox{object}{swim} at \colorbox{movie}{11am tomorrow}\\
\texttt{intent: add\_reminder}\\
\hline
\end{tabular}
\caption{English example from \name{} annotated with intents
(\texttt{add\_reminder}) and slots (\colorbox{object}{\texttt{todo}},
\colorbox{movie}{\texttt{datetime}}). The full set of languages is shown in
Table~\ref{fig:examples}.} 
\label{fig:sentence}
\end{figure}

Despite advances in neural modeling for slot and intent detection
(\S~\ref{sec:relmodels}), datasets for SLU remain limited, hampering progress
toward providing SLU for many language varieties. Most available datasets
either support only a specific domain (like air traffic
systems)~\cite{xu_end--end_2020}, or are broader but limited to English and a
few other languages~\cite{schuster-etal-2019-cross-lingual,coucke_snips_2018}.
We release \name, a new benchmark intended for SLU evaluation in low-resource
scenarios. \name{} contains evaluation data for 13 languages from six language
families, including a very low-resource dialect. It homogenizes annotation
styles of two recent
datasets~\cite{schuster-etal-2019-cross-lingual,coucke_snips_2018} and provides
the broadest public multilingual evaluation data for modern digital assistants.

\begin{table*}
    \centering
\small
    \begin{tabular}{l r r r r r r r}
    \toprule
    Dataset & Source & Langs. & Lang. Fams. & Domains\protect\footnotemark  & Intents & Slots & \#sents \\
    \midrule
    Atis & \newcite{hemphill1990atis} & 1 & 1 & 1 & 24 & 83 & 5,871 \\
    Snips & \newcite{coucke_snips_2018} & 1 & 1 & 5 & 7 & 39 & 14,484 \\
    HWU64 & \newcite{XLiu.etal:IWSDS2019} & 1 & 1 & 22 & 64 & 54 & 25,716 \\
    Almawave-SLU & \newcite{bellomaria2019almawave} & 1 & 1 & 1 & 7 & 39 & 8,542\\
    CSTOP$^\dagger$ & \newcite{einolghozati2021volumen} & 2 & 1 &  2 & 19 & 10 & 5,800 \\
    Leyzer$^*$ & \newcite{sowanski2020leyzer} & 3 & 1 & 20 & 186 & 86 & 16,257\\
    Facebook & \newcite{schuster-etal-2019-cross-lingual}& 3 & 2 &  3 & 12 & 11 & 57,049 \\
    multiAtis++ & \newcite{xu_end--end_2020} & 9 & 4 & 1 & 23 & 83 & 45,046 \\
    \midrule
    \name{} & This work & 13 & 6 & 7 & 16 & 33 & 10,000\\
    \bottomrule
    \end{tabular}
    \caption{Existing SLU datasets. Note that \name{} is intended to serve as
evaluation data only (Snips+Facebook can be used as English training data).
$^\dagger$Code-switched data (Spanglish). $^*$Automatically generated data.
Language families are counted based on highest level of
Glottolog~\cite{glottlog}.}

    \label{tab:datasets}
\end{table*}

Most previous efforts to multilingual SLU typically focus on translation or
multilingual embeddings transfer.  In this work, we propose an orthogonal
approach, and study non-English auxiliary tasks for transfer. We hypothesize
that jointly training on target language auxiliary tasks helps to learn
properties of the target language while learning a related task simultaneously.
We expect that this helps to refine the multilingual representations for better
SLU transfer to a new language. We evaluate a broad range of auxiliary tasks
not studied before in such combination, exploiting raw data, syntax in
Universal Dependencies (UD) and parallel data.

\paragraph{Our contributions} 
\begin{inparaenum}[i)]
    \item We provide \name, a new cross-lingual SLU evaluation dataset covering
Arabic (ar), Chinese (zh), Danish (da), Dutch (nl), English (en), German (de),
Indonesian (id), Italian (it), Japanese (ja), Kazakh (kk), Serbian (sr),
Turkish (tr) and an  Austro-Bavarian German dialect, South Tyrolean (de-st).\
    \item We experiment with new non-English auxiliary tasks for joint
cross-lingual transfer on slots and intents: UD parsing, machine translation
(MT), and masked language modeling.\
    \item We compare our proposed models to strong baselines, based on
multilingual pre-trained language models mBERT~\cite{devlin-etal-2019-bert} and
\texttt{xlm-mlm-tlm-xnli15-1024}~\cite{conneau-etal-2020-unsupervised}
(henceforth XLM15), where the former was pre-trained on 12 of our 13 languages,
and XLM15 on 5 of our 13 languages, thereby simulating a low-resource scenario.
We also compare to a strong machine translation
model~\cite{qin_multi-domain_2020}.
\end{inparaenum}

\footnotetext{The notion of domain is ill-defined within the scope of this
task. We report the numbers from the paper, and, for Snips, we have identified
the following: alarm, reminder, weather, restaurant, creative works.}

The remainder of this paper is structured as follows: we start by giving an
overview of existing datasets and introduce \name{} (\S~\ref{sec:sec2-data}),
then we discuss our baselines and proposed extensions (\S~\ref{sec:model}).
After this, we discuss the performance of these models (\S~\ref{sec:results}),
and provide an analysis (\S~\ref{sec:analysis}) before we end with the related
work on cross-lingual SLU (\S~\ref{sec:relmodels}) and the conclusion
(\S~\ref{sec:conclusion}).

%% file: data.tex
\subsection{Other SLU Datasets}\label{sec:datasets}
An overview of existing datasets is shown in Table~\ref{tab:datasets}. It
should be noted that we started the creation of \name{} at the end of 2019,
when less variety was available. We choose to use  the
Snips~\cite{coucke_snips_2018} and
Facebook~\cite{schuster-etal-2019-cross-lingual} data as a starting point.

Most existing datasets are English only (all datasets in
Table~\ref{tab:datasets} include English), and they differ in the domains they
cover. For example, Atis~\cite{hemphill1990atis} is focused on airline-related
queries, CSTOP~\cite{einolghozati2021volumen} contains queries about wheather
and devices, and other datasets cover multiple domains.

Extensions of Atis to new languages are a main direction. These include
translations to Chinese~\cite{he2013multi},
Italian~\cite{bellomaria2019almawave}, Hindi and
Turkish~\cite{upadhyay2018almost} and very recently, the MultiAtis++
corpus~\cite{xu_end--end_2020} with 9 languages in 4 language families. To the
best of our knowledge, this is the broadest publicly available SLU corpus to
date in terms of the number of languages, yet the data itself is less varied.
Almost simultaneously, \newcite{schuster-etal-2019-cross-lingual} provide a
dataset for three new topics (alarm, reminder, weather) in three languages
(English, Spanish and Thai). English utterances for a given intent were first
solicited from the crowd, translated into two languages (Spanish and Thai), and
manually annotated for slots.  We follow these approaches, but depart from the
Snips~\cite{coucke_snips_2018} and Facebook
\cite{schuster-etal-2019-cross-lingual} datasets to create a more varied
resource covering 13 languages, while homogenizing the annotations.

\label{sec:data}

\definecolor{movie}{HTML}{FF7F00}  
\definecolor{object}{HTML}{27ae60}  

\begin{table*}
\centering
\small
\resizebox{1.015\textwidth}{!}{ 
\hspace{-.35cm}
  \begin{tabular}{l l l }
  \toprule
  Lang. & Language Family & Annotation \\
  \midrule
  ar & Afro-Asiatic &
  \setcode{utf8}

\begin{tikzpicture}
    \path[use as bounding box] (0,0) rectangle (0,0);
        \node at (0.77,.075) [fill=object, minimum width=1.45cm, minimum height=.495cm] (v100) {};
\end{tikzpicture}

\<في دار السينما> \colorbox{movie}{Silly Movie 2.0}  \<أود أن أرى مواعيد عرض فيلم > 
\\

  da & Indo-European &
Jeg vil gerne se spilletiderne for \colorbox{movie}{Silly Movie 2.0} i \colorbox{object}{biografen} \\
    
    de & Indo-European &
Ich würde gerne den Vorstellungsbeginn für \colorbox{movie}{Silly Movie 2.0} im \colorbox{object}{Kino} sehen \\
    
    de-st & Indo-European &
 I mecht es Programm fir \colorbox{movie}{Silly Movie 2.0} in \colorbox{object}{Film Haus} sechn	\\

    en & Indo-European &
I'd like to see the showtimes for \colorbox{movie}{Silly Movie 2.0} at the \colorbox{object}{movie house} \\
    
    id & Austronesian &
Saya ingin melihat jam tayang untuk \colorbox{movie}{Silly Movie 2.0} di gedung \colorbox{object}{bioskop} \\
    
    it & Indo-European &
Mi piacerebbe vedere gli orari degli spettacoli per \colorbox{movie}{Silly Movie 2.0} al \colorbox{object}{cinema} \\

    ja & Japonic &
\begin{CJK}{UTF8}{min}
\colorbox{object}{映画館}の\colorbox{movie}{Silly Movie 2.0}の上映時間を見せて。
 \end{CJK}\\

   kk & Turkic & \selectlanguage{russian}
Мен \colorbox{movie}{Silly Movie 2.0} бағдарламасының \colorbox{object}{кинотеатрда} көрсетілім уақытын көргім келеді \\ 
    
    nl & Indo-European &
Ik wil graag de speeltijden van \colorbox{movie}{Silly Movie 2.0} in het \colorbox{object}{filmhuis} zien \\
    
    sr & Indo-European &
Želela bih da vidim raspored prikazivanja za \colorbox{movie}{Silly Movie 2.0} u \colorbox{object}{bioskopu} \\
    
    tr & Turkic &
\colorbox{movie}{Silly Movie 2.0'ın} \colorbox{object}{sinema salonundaki} seanslarını görmek istiyorum \\

    zh & Sino-Tibetan &
    \begin{CJK*}{UTF8}{bsmi}
我 想 看 \colorbox{movie}{Silly Movie 2.0} 在 \colorbox{object}{电影院} 的 放映 时间 
    \end{CJK*} \\
    \bottomrule
    \end{tabular}}
    \caption{Examples of annotation for all languages in our dataset with
intent: \texttt{SearchScreeningEvent}, and two slots:
\colorbox{movie}{\texttt{movie\_name}} and
\colorbox{object}{\texttt{object\_location\_type}}. Includes information on
language families from Glottolog~\cite{glottlog}.}
    \label{fig:examples}
\end{table*}

\name{} is a cross-lingual SLU evaluation dataset covering 13 languages from
six language families with English training data.  In what follows, we provide
details on the creation of \name{} (\S~\ref{sec:xslu}), including
homogenization of annotation guidelines and English source training data
(\S~\ref{sec:trainingdata}). For data statement and guidelines, we refer the
reader to Section~\ref{app:datastatement},~\ref{app:trans-guidelines}
and~\ref{app:ann-guidelines} in the Appendix.

\subsection{\name{}} \label{sec:xslu}
As a starting point, we extract 400 random English utterances from the Snips
data~\cite{coucke_snips_2018} as well as 400 from the Facebook
data~\cite{schuster-etal-2019-cross-lingual}, which for both consist of 250
utterances from the test-split and 150 from the dev-split. We maintain the
splits from the original data in \name{} (i.e.\ sentences in \name{} test are
from Snips test or Facebook test). We then translate this sample into all of
our target languages. It should be noted that some duplicates occur in the
random sample of the Facebook data. Since these instances naturally occur more
often, we decided to retain them to give a higher weight to common queries in
the final evaluation.\footnote{This decision has been made after discussion
with a real-world digital assistant team.}

\name{} includes Arabic (ar), Chinese (zh), Danish (da), Dutch (nl), English
(en), German (de), Indonesian (id), Italian (it), Japanese (ja), Kazakh (kk),
Serbian (sr), Turkish (tr) and an  Austro-Bavarian German dialect, South
Tyrolean (de-st).\footnote{The dialect is spoken by roughly 450,000 speakers in
an Alpine province in Northern Italy. It has no official ISO language code nor
a normed writing form.} We have 13 evaluation languages with 800 sentences per
language\footnote{Except for Japanese where we only have the Facebook data.}
resulting in a final dataset of 10,000 sentences. The language selection is
based on availability of translators/annotators (most of them are co-authors of
this paper, i.e.\ highly-educated with a background in NLP). We favor this
setup over crowd-sourcing, i.e.\ quality and breadth in annotation and
languages, and because for some languages crowd-sourcing is not an
option.\footnote{We did not have access to native speakers of Thai and Spanish
part of the Facebook data~\cite{schuster-etal-2019-cross-lingual}, which is why
there are not included (yet).} For more information on the data and annotators
we refer to the dataset statement in Appendix~\ref{app:datastatement}.

The first step of the dataset creation was the translation. For this, the goal
was to provide a fluent translation which was as close as possible to the
original meaning. Because the data consists of simple, short utterances, we
consider our annotator pool to be adequate for this task (even though they are
not professional translators). The intents could easily be transferred from the
English data, but the slots needed to be re-annotated, which was done by the
same annotators.

Unfortunately, we were unable to retrieve annotation guidelines from the
earlier efforts. Hence, as a first step of and as part of training, we derived
annotation guidelines by jointly re-annotating dev and test portions of the
English parts of the two data sources. These guidelines were revised multiple
times in the process to derive the final guidelines for the whole dataset.
Ultimately,  the data collection process proceeded in two steps: translation of
the data from English, and slot annotation in the target language. The aim of
the guidelines was to generalize labels to make them more broadly applicable to
other intent subtypes, and remove within-corpus annotation variation (see
Appendix~\ref{app:ann-guidelines} for details).  We calculated inter-annotator
agreement for the  guidelines; three annotators native in Dutch annotated 100
samples, and reached a Fleiss Kappa~\cite{fleiss1971measuring} score of 0.924,
which is very high agreement. Common mistakes included annotation of question
words, inclusion of locations in reminders, and the inclusion of function words
in the spans. We updated the guidelines after the agreement study. After these
target phase annotation rounds, we finalized the guidelines, which are provided
in the Appendix~\ref{app:ann-guidelines} and form the basis for the provided
data.  Table~\ref{fig:examples} provides an example annotation for all 13
languages for the example sentence ``I'd like to see the showtimes for Silly
Movie 2.0 at the movie house''. These example translations illustrate not only
the differences in scripts, but also differences in word order and length of
spans, confirming the distances between the languages.

\subsection{English Training Data}\label{sec:trainingdata}
Because of our revised guidelines for the Facebook data and mismatches in
granularity of labels between the Snips and Facebook data, we  homogenize the
original training data for both sources and include it in our release. For the
Facebook data, this includes rule-based fixing of spans and recognition of the
\textsc{reference} and \textsc{recurring\_time} labels.\footnote{For more
details on this procedure, we refer to \texttt{scripts/0.fixOrigAnnotation.py}
in the repo.} For the Snips data, we convert a variety of labels that describe
a location to the \textsc{location} label which is used in the Facebook data,
and labels describing a point or range in time to \textsc{datetime}. After this
process, we simply concatenate both resulting datasets, and shuffle them before
training. The resulting training data has 43,605 sentences.

%% file: models.tex
Our main hypothesis is that we can improve zero-shot transfer with
target-language auxiliary tasks. We hypothesize that this will help the
multilingual pre-trained base model to learn peculiarities about the target
language, while it is learning the target task as well. To this end, we use
three (sets of) tasks with a varying degree of complexity and availability: 1)
Masked Language Modeling (MLM): which is in spirit similar to pre-training on
another domain~\cite{gururangan-etal-2020-dont}, however, we learn this jointly
with the target task to avoid catastrophic
forgetting~\cite{mccloskey1989catastrophic}; 2) Neural Machine Translation
(NMT): where we learn English SLU as well as translation from English to the
target language; and 3) Universal Dependency (UD) parsing: to insert linguistic
knowledge into the shared parameter space to learn from syntax as auxiliary
task besides learning the SLU task. 

In the following subsections, we first describe the implementation of our
baseline model, and the machine translation-based model, and then describe the
implementation of all auxiliary tasks (and the data used to train them).
Auxiliary tasks are sorted by dataset availability (MLM $\succ$ NMT $\succ$
UD), where the first type can be used with any raw text, the second one needs
parallel data -- which is readily available for many languages as a by-product
of multilingual data sources -- and the last one requires explicit human
annotation. For South Tyrolean, a German dialect, no labeled target data of any
sort is available; we use the German task data instead. We provide more details
of data sources and sizes in Appendix~\ref{app:aux-datasets}.

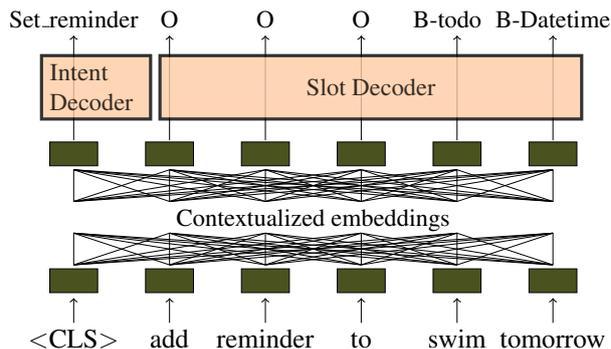
\begin{figure}
    \resizebox{\columnwidth}{!}{
    \input{imgs/model}}
    \caption{Overview of the baseline model.}
    \label{fig:model}
\end{figure}

\subsection{Baseline}
All our models are implemented in MaChAmp
v0.2~\cite{vandergoot-etal-2020-machamp}, an
AllenNLP-based~\cite{Gardner2017AllenNLP} multi-task learning toolkit. It uses
contextual embeddings, and fine-tunes them during training. In the multi-task
setup, the encoding is shared, and each task has its own decoder. For slot
prediction, a greedy decoding with a softmax layer is used, for intents it uses
a linear classification layer over the \texttt{[CLS]} token (see
Figure~\ref{fig:model}).\footnote{ We also tried to use a CRF layer for slots
which consistently led to lower performance.} The data for each task is split
in batches, and the batches are then shuffled.  We use the default
hyperparameters of MaChAmp for all experiments which were optimized on a wide
variety of tasks~\cite{vandergoot-etal-2020-machamp}.\footnote{Hyperparameter
settings used in experiments are reported in Appendix~\ref{app:hyperparams}.}
The following models are extensions of this baseline. In the NMT-transfer model
(\S~\ref{sec:nmt-transfer}), the training data is translated before passing it
into the model. For the auxiliary models (\S~\ref{sec:aux-mlm},
\ref{sec:aux-nmt} and \ref{sec:aux-ud}), we simply add another decoder next to
the intent and slot decoders. The losses are summed, and typically weighted
(multiplied) by a factor which is given in corresponding subsections.  We
enable the proportional sampling option of MaChAmp (multinomial sampling
$\alpha=0.5$) in all multi-task experiments, to avoid overfitting to the
auxiliary task.

\subsection{Neural Machine Translation with Attention (nmt-transfer)} \label{sec:nmt-transfer}
For comparison, we trained a NMT model to translate the NLU training data into
the target language, and map the annotations using attention. As opposed to
most previous work using this
method~\cite{xu_end--end_2020,he2013multi,schuster-etal-2019-cross-lingual}, we
opt for an open-source implementation and provide the scripts to rerun the
experiments. More specifically, we use the Fairseq toolkit
~\cite{ott2019fairseq} implementation of the Transformer-based model
\cite{vaswani2017attention} with default hyperparameters.  Sentences were
encoded using byte-pair encoding (BPE) \cite{sennrich-etal-2016-neural}, with a
shared vocabulary of 32,000 tokens.  At inference time, we set the beam size to
4, and extracted alignment scores to target tokens calculated from the
attention weights matrix.  These scores are used to align annotation labels to
target language outputs; we map the label of each token to the highest scoring
alignment target token. We convert the output to valid BIO tags: we use the
label of the B for the whole span, and an I following an O is converted to a B.
\paragraph{Data}
To ensure that our machine translation data is suitable for the target domain,
we choose to use a combination of transcribed spoken parallel data.  For
languages included in the IWSLT 2016 Ted talks dataset~\cite{cettolo2016iwslt},
we use the train and development data included, and enlarge the training data
with the training split from
Opensubtitles\footnote{http://www.opensubtitles.org/} 2018
~\cite{lison-tiedemann-2016-opensubtitles2016}, and
Tatoeba~\cite{tiedemann-2012-parallel}. For languages absent in IWSLT2016, we
used the Opensubtitles data for training and Tatoeba as development set.  For
Kazakh, the Opensubtitles data only contains $~$2,000 sentences, so we
concatenated out-of-domain data from the WMT2019
data~\cite{barrault-etal-2019-findings}, consisting of English-Kazakh crawled
corpora.  We adapt the BertBasic tokenizer (which splits punctuation, it does
not perform subword tokenization) to match the Facebook and Snips dataset
tokenization and use this to pre-tokenize the data.

\subsection{Masked Language Modeling (aux-mlm)} \label{sec:aux-mlm}
Previous work has shown that continuing to train a language model with an MLM
objective on raw data close to the target domain leads to performance
improvements~\cite{gururangan-etal-2020-dont}. However, in our setup,
task-specific training data and target data are from different languages.
Therefore, in order to learn to combine the language and the task in a
cross-lingual way, we train the model jointly with MLM and task-specific
classification objective on target and training languages respectively. We
apply the original BERT masking strategy and we do not include next sentence
prediction following~\newcite{liu2019roberta}. For computational efficiency, we
limit the number of input sentences to 100,000 and use a loss weight of 0.01
for MLM training.
\paragraph{Data}
For our masked language modeling objective, we use the target language machine
translation data described above.

\input{mainTable}

\subsection{Machine Translation (aux-nmt)} \label{sec:aux-nmt}
To jointly learn to transfer linguistic knowledge from English to the target
language together with the target task, we implement a NMT decoder based on the
shared encoder. We use a sequence-to-sequence
model~\cite{sutskever2014sequence} with a recurrent neural network decoder,
which suits the auto-regressive nature of the machine translation
tasks~\cite{cho-etal-2014-learning}, and an attention mechanism to avoid
compressing the whole source sentence into a fixed-length
vector~\cite{bahdanau2014neural}. We found that fine-tuning the shared encoder
achieves good performance on our machine translation
datasets~\cite{conneau2019cross,clinchant-etal-2019-use}, alleviating the need
for freezing its parameters during training in order to avoid catastrophic
forgetting~\cite{imamura-sumita-2019-recycling,goodfellow2013empirical}.
Similar to MLM, we use 100,000 sentences, and a weight of 0.01.
\paragraph{Data} 
For this auxiliary task, we use the same data as for \textsc{nmt-transfer},
described in detail above.

\subsection{Universal Dependencies (aux-ud)} \label{sec:aux-ud}
Using syntax in hierarchical multi-task learning has previously shown to be
beneficial~\cite{hashimoto-etal-2017-joint,godwin2016deep}. We here use full
Universal Dependency (UD) parsing, i.e., part-of-speech (POS) tagging,
lemmatization, morphological tagging and dependency parsing as joint auxiliary
tasks, as opposed to previous hierarchical MTL work.  For all tasks we use the
default settings of MaChAmp and set the loss weight of each  UD subtask to
0.25.
\paragraph{Data} 
For each language, we manually picked a matching UD treebank from version
2.6~\cite{nivre-etal-2020-universal} (details in the Appendix). Whenever
available, we picked an in-language treebank, otherwise we choose a related
language. We used size, annotation quality, and domain as criteria.

%% file: imgs/model.tex
\def\wordpieces#1#2#3#4#5#6#7
{
\begin{scope}
\newcount\wordPiece
\wordPiece=0
\loop
    \node (#7\the\wordPiece) [draw, minimum width=2*#4cm, minimum height=#4cm, fill=#6] at (#1 + \the\wordPiece * 4 * #4, #2) {};
    \advance \wordPiece +1
\ifnum \wordPiece<#3
\repeat
\end{scope}
}

\def\arrows#1#2#3#4#5#6
{
\begin{scope}
\newcount\wordCounterOne
\wordCounterOne=0
\loop
    \draw [-, opacity=#6] (#1 + \wordCounterOne * 4 * #3, #2) -- (#1 + 0 * 4 * #3, #2+#5);
    \draw [-, opacity=#6] (#1 + \wordCounterOne * 4 * #3, #2) -- (#1 + 1 * 4 * #3, #2+#5);
    \draw [-, opacity=#6] (#1 + \wordCounterOne * 4 * #3, #2) -- (#1 + 2 * 4 * #3, #2+#5);
    \draw [-, opacity=#6] (#1 + \wordCounterOne * 4 * #3, #2) -- (#1 + 3 * 4 * #3, #2+#5);
    \draw [-, opacity=#6] (#1 + \wordCounterOne * 4 * #3, #2) -- (#1 + 4 * 4 * #3, #2+#5);
    \draw [-, opacity=#6] (#1 + \wordCounterOne * 4 * #3, #2) -- (#1 + 5 * 4 * #3, #2+#5);
    \advance \wordCounterOne +1
\ifnum \wordCounterOne<#4
\repeat
\end{scope}
}

\definecolor{armygreen}{rgb}{0.29, 0.32, 0.12	}
\definecolor{brickred}{rgb}{0.8, 0.25, 0.33}
\definecolor{darksalmon}{rgb}{0.91, 0.59, 0.48}
\definecolor{deeppeach}{rgb}{1.0, 0.8, 0.64}
\definecolor{deepchampagne}{rgb}{0.98, 0.84, 0.65}
\definecolor{darkgreen}{rgb}{0.0, 0.2, 0.13}
\definecolor{airforceblue}{rgb}{0.36, 0.54, 0.66}
\begin{tikzpicture}
    \path[use as bounding box] (.15,1) rectangle (8.25,6.7);

\wordpieces{0}{4.25cm}{6}{.375}{1}{armygreen}{wordEnc}
\wordpieces{0}{2.25cm}{6}{.375}{1}{armygreen}{wordEnc}
\node (word1) [minimum height=.25cm, text height=.25cm,text depth=.25cm] at (0,  1.2) {\large $<$CLS$>$};
\node (word2) [minimum height=.25cm, text height=.25cm,text depth=.25cm] at (1.5,1.2) {\large add};
\node (word3) [minimum height=.25cm, text height=.25cm,text depth=.25cm] at (3,  1.2) {\large reminder};
\node (word4) [minimum height=.25cm, text height=.25cm,text depth=.25cm] at (4.5,1.2) {\large to};
\node (word5) [minimum height=.25cm, text height=.25cm,text depth=.25cm] at (6,  1.2) {\large swim};
\node (word6) [minimum height=.25cm, text height=.25cm,text depth=.25cm] at (7.5,1.2) {\large tomorrow};
\arrows{0}{4}{.375}{6}{-.5}{1}

\arrows{0}{2.5}{.375}{6}{.5}{1}

\node (Bert) [rectangle, line width=.05cm, minimum width=8.5cm, minimum height=1.5cm] at (3.75,3.2) {Contextualized embeddings};

\node (feats1) [rectangle, line width=.05cm, armygreen, minimum width=.85cm, minimum height=3.5cm] at (0,3.75) {};
\node (feats2) [rectangle, line width=.05cm, armygreen, minimum width=.85cm, minimum height=3.5cm] at (1.5,3.75) {};
\node (feats3) [rectangle, line width=.05cm, armygreen, minimum width=.85cm, minimum height=3.5cm] at (3,3.75) {};
\node (feats4) [rectangle, line width=.05cm, armygreen, minimum width=.85cm, minimum height=3.5cm] at (4.5,3.75) {};
\node (feats5) [rectangle, line width=.05cm, armygreen, minimum width=.85cm, minimum height=3.5cm] at (6,3.75) {};
\node (feats6) [rectangle, line width=.05cm, armygreen, minimum width=.85cm, minimum height=3.5cm] at (7.5,3.75) {};

\node (sentOut) [minimum height=.25cm, text height=.25cm,text depth=.25cm] at (0,6.25) {\normalsize Set\_reminder};
\node (upos1) [minimum height=.25cm, text height=.25cm,text depth=.25cm] at (1.5,6.25) {\normalsize O};
\node (upos2) [minimum height=.25cm, text height=.25cm,text depth=.25cm] at (3,6.25) {\normalsize O};
\node (upos3) [minimum height=.25cm, text height=.25cm,text depth=.25cm] at (4.5, 6.25) {\normalsize O};
\node (upos4) [minimum height=.25cm, text height=.25cm,text depth=.25cm] at (6,6.25) {};
\node (upos4word) [minimum height=.25cm, text height=.25cm,text depth=.25cm] at (5.85,6.25) {\normalsize B-todo};
\node (upos5) [minimum height=.25cm, text height=.25cm,text depth=.25cm] at (7.5,6.25) {\normalsize B-Datetime};

\draw [->]  ([yshift=-1.4cm]sentOut.south) -- ([yshift=.25cm]sentOut.south);
\draw [->] ([yshift=-1.4cm]upos1.south) -- ([yshift=.25cm]upos1.south);
\draw [->] ([yshift=-1.4cm]upos2.south) -- ([yshift=.25cm]upos2.south);
\draw [->] ([yshift=-1.4cm]upos3.south) -- ([yshift=.25cm]upos3.south);
\draw [->] ([yshift=-1.4cm]upos4.south) -- ([yshift=.25cm]upos4.south);
\draw [->] ([yshift=-1.4cm]upos5.south) -- ([yshift=.25cm]upos5.south);

\draw [->] (word1) -- (feats1);
\draw [->] (word2) -- (feats2);
\draw [->] (word3) -- (feats3);
\draw [->] (word4) -- (feats4);
\draw [->] (word5) -- (feats5);
\draw [->] (word6) -- (feats6);


\node (polDecoder) [rectangle, line width=.05cm, draw, fill=deeppeach, text width=1.45cm, opacity=.8, minimum width=1.45cm, minimum height=1cm] at (0.35,5.3) {Intent Decoder};
\node (uposDecoder) [rectangle, line width=.05cm, draw, fill=deeppeach, opacity=.8, minimum width=6.6cm, minimum height=1cm] at (4.65,5.3) {Slot Decoder};


\end{tikzpicture}

%% file: mainTable.tex
\begin{table*}[ht!]
\centering
\small
\setlength{\tabcolsep}{5pt}
\def\arraystretch{.9}
\begin{tabular}{l| r | r r r r r r r r r r r r|r}
\toprule
mBERT & \textbf{en} & de-st & \textbf{de} & \textbf{da} & \textbf{nl} & \textbf{it} & \textbf{sr} & \textbf{id} & \textbf{ar} & \textbf{zh} & \textbf{kk} & \textbf{tr} & \textbf{ja}$^*$ & Avg.\\
lang2vec & --- & --- & 0.18 & 0.18 & 0.19 & 0.22 & 0.23 & 0.24 & 0.30 & 0.33 & 0.37 & 0.38 & 0.41 \\
\midrule
Slots\\
\midrule
base & \textbf{97.6} & 48.5 & 33.0 & 73.9 & 80.4 & 75.0 & \textbf{67.4} & \textbf{71.1} & 45.8 & \textbf{72.9} & 48.5 & 55.7 & 59.9 & 61.0 \\
nmt-transfer & 0.0 & 50.9 & 34.5 & 60.8 & 63.7 & 51.0 & 41.3 & 54.2 & \textbf{48.2} & 27.9 & 0.2 & 52.0 & 45.0 & 44.1 \\
aux-mlm & 97.3 & \textbf{53.0} & \textbf{34.6} & \textbf{75.9} & \textbf{82.2} & \textbf{78.0} & 63.8 & 69.5 & 48.1 & 69.4 & \textbf{51.3} & \textbf{58.4} & \textbf{63.5} & \textbf{62.3} \\
aux-nmt & 0.0 & 44.5 & 33.3 & 71.4 & 76.9 & 71.9 & 58.5 & 62.9 & 38.7 & 70.3 & 38.2 & 50.2 & 58.7 & 56.3 \\
aux-ud & 97.5 & 47.6 & 29.1 & 73.7 & 73.3 & 61.8 & 56.8 & 61.1 & 42.6 & 64.9 & 45.2 & 53.8 & 47.6 & 54.8 \\
\midrule
Intents\\
\midrule
base & \textbf{99.7} & 67.8 & 74.2 & 87.5 & 72.3 & 81.7 & 75.7 & 80.7 & 63.1 & 83.3 & \textbf{60.1} & 74.7 & 53.9 & 72.9 \\
nmt-transfer & 0.0 & \textbf{89.9} & \textbf{97.5} & \textbf{97.3} & \textbf{98.3} & \textbf{96.8} & \textbf{92.5} & \textbf{98.1} & \textbf{89.2} & \textbf{97.3} & 24.6 & \textbf{98.3} & \textbf{78.8} & \textbf{88.2} \\
aux-mlm & 99.5 & 66.3 & 75.4 & 80.7 & 73.5 & 80.1 & 65.4 & 72.2 & 59.7 & 78.3 & 47.6 & 62.2 & 42.9 & 67.0 \\
aux-nmt & 0.0 & 63.0 & 72.8 & 86.6 & 70.8 & 78.7 & 71.9 & 75.5 & 56.4 & 80.9 & 56.3 & 68.9 & 53.6 & 69.6 \\
aux-ud & 99.5 & 62.6 & 58.2 & 67.7 & 60.1 & 62.9 & 59.9 & 65.0 & 45.7 & 70.4 & 24.6 & 43.9 & 38.5 & 55.0 \\
\midrule[1pt]
XLM15 & \textbf{en} & de-st & \textbf{de} & da & nl & it & sr & id & \textbf{ar} & \textbf{zh} & kk & \textbf{tr} & ja$^*$ \\
\midrule
Slots\\
\midrule
base & 97.0 & 39.4 & 33.3 & 26.3 & 30.9 & 27.3 & 15.9 & 14.9 & \textbf{49.1} & 57.6 & 10.9 & 45.5 & 33.4 & 32.1 \\
nmt-transfer & 0.0 & 43.0 & 34.7 & \textbf{59.5} & \textbf{61.9} & \textbf{49.4} & \textbf{39.7} & \textbf{53.0} & 47.2 & 28.4 & 0.0 & 50.5 & \textbf{34.7} & \textbf{41.8} \\
aux-mlm & \textbf{97.2} & 44.0 & \textbf{35.9} & 45.8 & 49.5 & 40.7 & 18.7 & 24.6 & 48.9 & \textbf{64.8} & \textbf{13.6} & \textbf{60.5} & 30.4 & 39.8 \\
aux-nmt & 0.0 & 32.3 & 32.3 & 26.3 & 28.0 & 24.1 & 12.4 & 13.7 & 38.0 & 29.4 & 7.2 & 33.1 & 16.6 & 24.4 \\
aux-ud & 97.0 & \textbf{46.0} & 34.6 & 36.3 & 45.4 & 45.3 & 22.0 & 21.6 & 45.1 & 52.5 & 13.1 & 50.1 & 33.2 & 37.1 \\
\midrule
Intents\\
\midrule
base & \textbf{99.7} & 61.3 & 78.5 & 56.3 & 45.4 & 48.0 & 41.4 & 36.4 & 67.5 & 78.8 & 29.9 & 67.3 & 39.1 & 54.1 \\
nmt-transfer & 0.0 & \textbf{80.6} & \textbf{97.6} & \textbf{97.6} & \textbf{97.7} & \textbf{96.6} & \textbf{92.4} & \textbf{96.7} & \textbf{88.4} & \textbf{97.1} & 16.7 & \textbf{97.9} & \textbf{61.7} & \textbf{85.1} \\
aux-mlm & 99.6 & 63.9 & 86.3 & 62.8 & 59.9 & 53.0 & 31.4 & 42.1 & 64.0 & 86.5 & 25.7 & 63.3 & 44.4 & 56.9 \\
aux-nmt & 0.0 & 52.0 & 52.0 & 60.7 & 44.7 & 44.7 & 40.3 & 43.1 & 54.3 & 54.1 & 21.0 & 53.2 & 25.2 & 45.4 \\
aux-ud & 99.5 & 62.6 & 72.2 & 47.2 & 42.3 & 52.5 & 33.6 & 31.8 & 45.7 & 57.1 & \textbf{30.3} & 51.1 & 35.5 & 46.8 \\
\bottomrule

\end{tabular}
\caption{Results on slot labeling (in strict F1) and intent classification (in
accuracy) on the development split of all 13 languages. Average over 5 seeds,
standard deviations can be found in Appendix~\ref{sec:stdev}. Sorted by
language distance to \textit{en} (de-st excluded), which is the cosine distance
between the syntax, phonology and inventory vectors of
lang2vec~\cite{littell-etal-2017-uriel}. \textbf{Bold} languages are included
during pre-training. The last column (Avg.)~is the average of all cross-lingual
experiments, i.e.\ without English. $^*$For Japanese we have 50\% less
evaluation data.} 
\label{tab:mainResults}
\end{table*}

%% file: results.tex
\begin{figure*}
    \centering
    \begin{subfigure}[b]{0.495\textwidth}
         \centering
         \includegraphics[width=.95\textwidth]{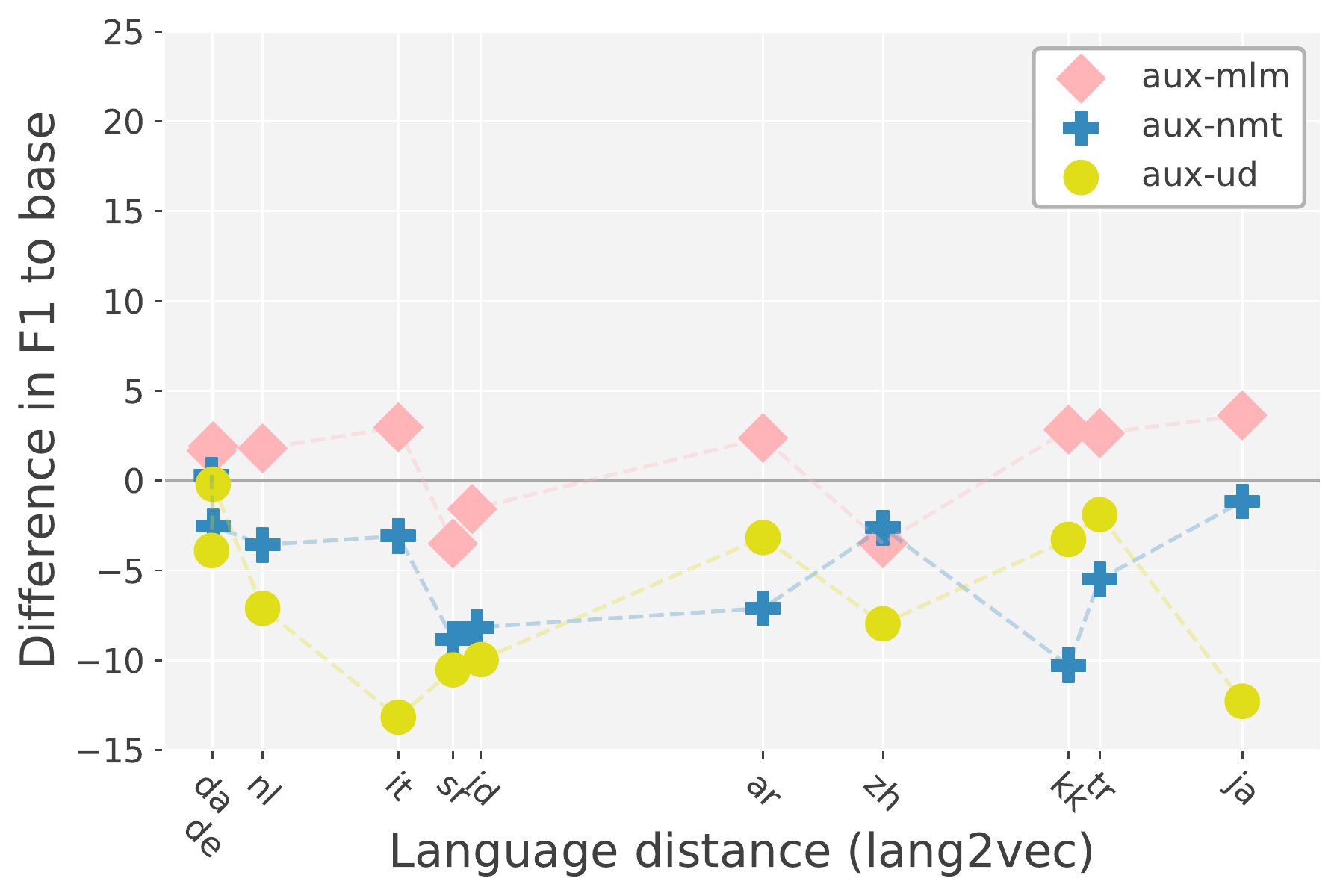}
         \caption{}
         \label{fig:langDist-mbert}
     \end{subfigure}
     \hfill
     \begin{subfigure}[b]{0.495\textwidth}
        \includegraphics[width=.95\textwidth]{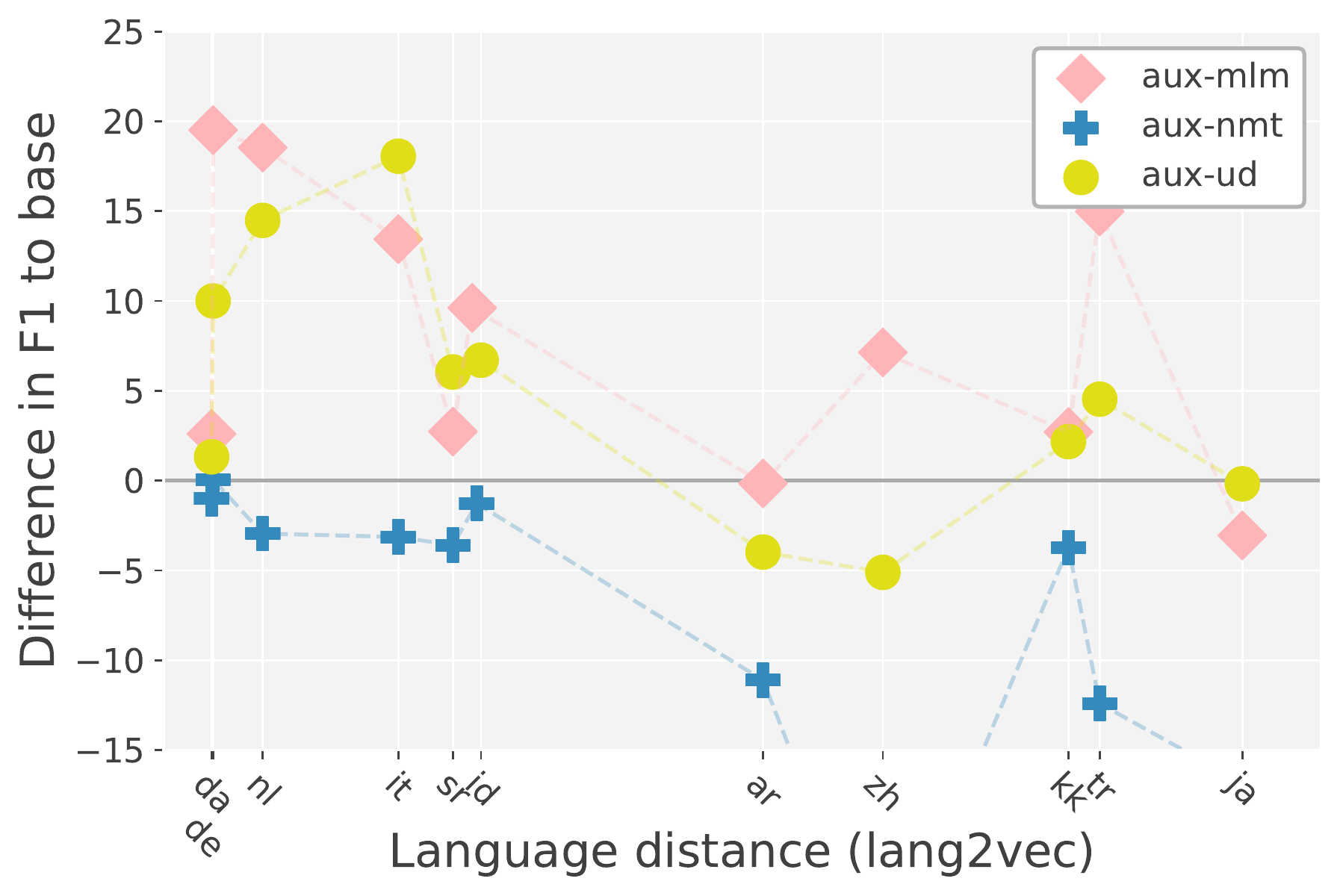}
        \caption{}
        \label{fig:langDist-xlm}
    \end{subfigure}
    \caption{Performance increase over baseline for each auxiliary task
with respect to the language distance (lang2vec) to English for mBERT (a) and
XLM15 (b). It should be noted that the lines carry no meaning (i.e. we can not
conclude performance based on language distance alone), and are shown to make
trends visible.}
\end{figure*}

\subsection{Experimental Setup}
We target a low-resource setup, and hence all our experiments assume
\textit{no} target-language training nor development data for the target task.
For all our experiments we use the English training from the Facebook  and
Snips data, and their English development sets (all converted to match our
guidelines, see \S~\ref{sec:sec2-data}). We use strict-span F1 score for slots
(where both span and label must match exactly) and accuracy for intents as main
evaluation metric as is standard for these tasks.\footnote{Ill-formed spans are
automatically converted to match the BIO-scheme (first word with I is converted
to B, and B-I spans with different labels are converted to all match the first
label).}  All reported results (including analysis and test data) are the
average over 5 runs with different random seeds.

To choose the final model, we use the scores on the English development data.
We are aware that this was recently shown to be sub-optimal in some
settings~\cite{keung-etal-2020-dont}, however there is no clear solution on how
to circumvent this in a pure zero-shot cross-lingual setup (i.e.\ without
assuming any target language target task annotation data). 

We use multilingual BERT (mBERT) as contextual encoder for our experiments. We
are also interested in low-resource setups. As all of our languages are
included in pre-training of mBERT (except the de-st dialect), we also study
XLM15 (\textsc{xlm-mlm-tlm-xnli15-1024}), which in pre-training covers only 5
of the 13 \name{} languages, to simulate further a real low-resource setup.

Table~\ref{tab:mainResults} reports the scores on 13 \name{} languages, for 2
tasks (slot and intent prediction) and 2 pre-trained language models. Languages
are ordered by language distance, whenever available. Below we discuss the main
findings per task.

\paragraph{Slots}
For slot filling, auxiliary tasks are beneficial for the majority of the
languages, and the best performing multi-task model (aux-mlm) achieves
\texttt{+1.3} for mBERT and \texttt{+7.7} for XLM15 average improvements over
the baseline. By comparing mBERT and XLM15, there are significant performance
drops for languages not seen during XLM15 pre-training, e.g., Danish (da) and
Indonesian (id). This confirms that having a language in pre-training has a
large impact on cross-lingual transfer for this task.  For other languages
involved in pre-training, both aux-mlm and aux-ud beat the baseline model. This
supports our hypothesis that, after multilingual pre-training, auxiliary tasks
(with token-level prediction both self-supervised and supervised) help the
model learn the target language and a better latent alignment for cross-lingual
slot filling.

\paragraph{Intents}
For intent classification the nmt-transfer model is very strong as it uses
explicit translations, especially for languages not seen during pre-training.
Using nmt as an auxiliary task does not come close, however, it should be noted
that this only uses a fraction of the data and computational costs (see
\S~\ref{sec:costs}).  One main limitation of the nmt-transfer model is that it
is dependant on a high-quality translation model, which in turn requires a
large quantity of in-domain parallel data. Results on Kazakh (kk) confirm this,
where the translation model is trained on out-of-domain data, because in-domain
data was not available (\S~\ref{sec:nmt-transfer}).

\begin{table}
\centering
        \resizebox{.95\columnwidth}{!}{
    \begin{tabular}{l | l l | l l}
\toprule
 & \multicolumn{2}{c}{\name{}} & \multicolumn{2}{c}{MultiAtis++} \\
Model & Slots & Intents & Slots & Intents\\
\midrule
mBERT &&&& \\
\midrule
base & 61.00 & 72.91 & 71.12 & 87.28 \\ 
nmt-transfer & 44.13$^3$ & 88.22$^{11}$ & 49.20$^1$ & 92.82$^8$ \\ 
aux-mlm & 62.32$^9$ & 67.02$^1$ & 69.15$^1$ & 83.33$^0$ \\ 
aux-nmt & 56.29$^0$ & 69.62$^0$ & 66.95$^1$ & 84.28$^0$ \\ 
aux-ud & 54.81$^0$ & 54.97$^0$ & 54.18$^1$ & 64.81$^0$ \\ 
\midrule
XLM15 &&&&\\
\midrule
base & 32.05 & 54.15 & 22.71 & 70.63 \\ 
nmt-transfer & 41.85$^8$ & 85.08$^{11}$ & 20.97$^4$ & 83.57$^8$ \\ 
aux-mlm & 39.77$^{10}$ & 56.94$^8$ & 62.10$^8$ & 81.54$^7$ \\ 
aux-nmt & 23.85$^0$ & 43.68$^2$ & \hspace{.2cm}0.31$^0$ & 42.12$^0$ \\ 
aux-ud & 37.10$^9$ & 46.82$^1$ & 52.95$^6$ & 80.30$^7$ \\ 
\bottomrule
    \end{tabular}}
    \caption{Results on the test data, average over all languages except
English. Significance tested with almost stochastic order~\cite{dror2019deep}
test with Bonferroni correction~\cite{bonferroni1936teoria} as implemented
by~\newcite{dennis_ulmer_2021_4638709}: 1,000 iterations p=0.05. The number in
superscript indicates the number of languages (/12) with significant
improvements compared to the baseline.}
    \label{tab:test}
\end{table}

\subsection{Test Data}
Our main findings are confirmed on the test data (Table~\ref{tab:test}), where
we also evaluate on MultiAtis++. The nmt-transfer model perform superior on
intents, whereas its performance on slots is worse. The best auxiliary setups
are aux-mlm followed by aux-ud. Most significant gains with auxiliary tasks are
obtained for languages not included in pre-training (XLM15). We believe there
is a bug for aux-nmt with XLM15 (see 
also results in Appendix~\ref{app:aux-tasks}), which we unfortunately could not 
resolve before submission time. Furthermore, we believe more tuning of machine 
translation can increase its viability as auxiliary task. In general our results 
on MultiAtis++ are lower compared to~\newcite{xu_end--end_2020}, which is 
probably because they used a black-box translation model.

%% file: analysis.tex
\subsection{Effect of Language Distance}
In Figure~\ref{fig:langDist-mbert} we plot the performance increase over
baseline for each auxiliary task with respect to the language distance when
using mBERT. The results confirm that aux-mlm is the most promising auxiliary
model, and clearly show that it is most beneficial for languages with a large
distance to English. Figure~\ref{fig:langDist-xlm} shows the same plot for the
XLM15 models, and here the trends are quite different. First, we see that also
for close languages, aux-ud as well as aux-mlm are beneficial. Second, the
aux-ud model also performs better for the more distant languages.

\begin{figure}
    \centering
    \includegraphics[width=\columnwidth]{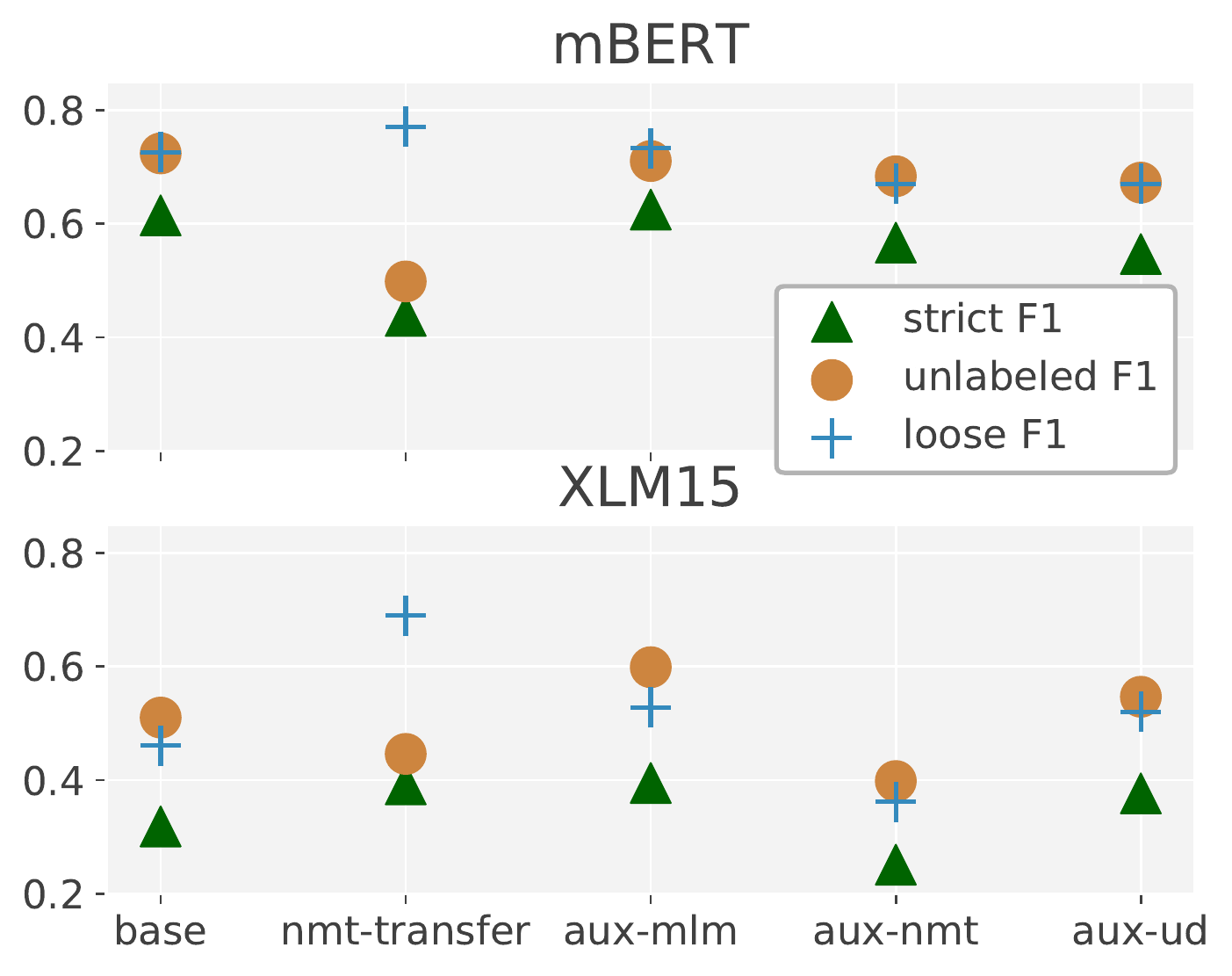}
    \caption{F1 scores variants for each model, averaged over 12 languages
(English is not included).}
    \label{fig:f1s}
\end{figure}

\subsection{Slot Detection Versus Classification}
To evaluate whether the detection of the slots or the classification of the
label is the bottleneck, we experiment with two varieties of the F1 score.  For
the first variant, we ignore the label and consider only whether the span is
correct. We refer to this as unlabeled F1.  For span detection, we allow for
partial matches (but with the same label) which count towards true positives
for precision and recall. We refer to this metric as loose F1. 

Average scores with all three F1 scores for both pre-trained embeddings are
plotted in Figure~\ref{fig:f1s}. One of the main findings is that nmt-transfer
does very well on the loose F1 metric, which means that it is poor at finding
spans, instead of labeling them. For the other models the difference between
strict and unlabeled F1 is smaller, and both can gain approximately 5-10\%
absolute score for both types of errors. The only other large difference is for
aux-nmt with XLM15, which makes more errors in the labeling (unlabeled F1 is
higher). An analysis of the per-language results show that this is mainly due
to errors made in the Kazakh dataset.

\subsection{Correlation Auxiliary Task Performance}
\begin{figure}
    \includegraphics[width=\columnwidth]{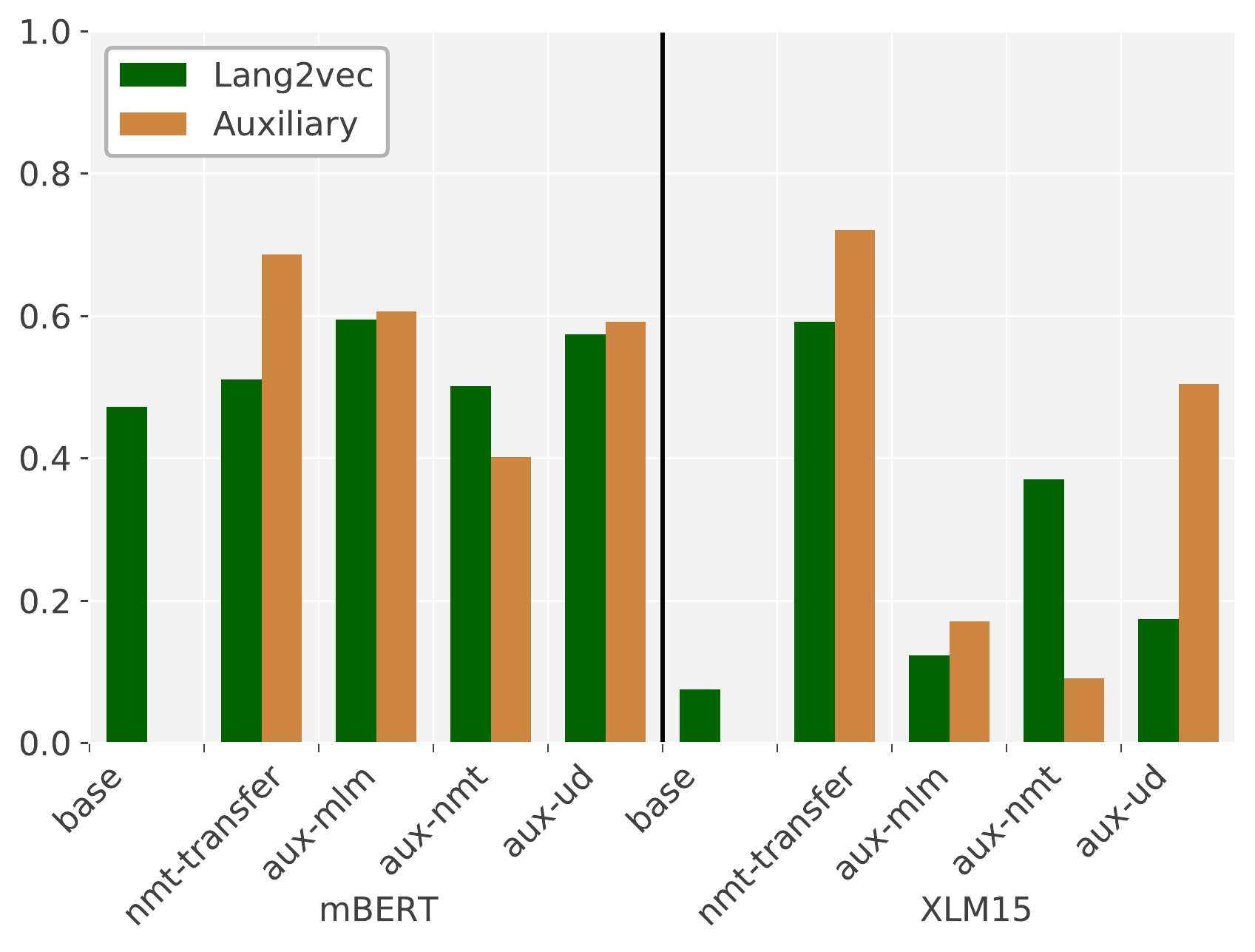}
    \caption{Pearson correlations between target tasks performance  (average of
slots/intents) and 1) language distance as estimated by lang2vec, and 2) the
auxiliary task. For nmt-transfer, the auxiliary task is the BLEU score of the
machine translation, and for the baseline there is no auxiliary task.}
    \label{fig:correlations}
\end{figure}

In Figure~\ref{fig:correlations} we plot the absolute Pearson correlations
between the auxiliary task (auxiliary task performance can be found in
Appendix~\ref{app:aux-tasks}) and the target tasks performance as well as
between the target tasks and the language distance (from lang2vec, see
Table~\ref{tab:mainResults}). Here we use the average of slots/intents as score
for the target task. The results show that when using only datasets from
languages included in the pre-trained language model (i.e., mBERT), both
language distance and auxiliary task performance are competitive predictors,
whereas if also new languages are considered (XLM15) auxiliary task performance
is clearly a stronger predictor. 

\subsection{Computational Costs}
\label{sec:costs}
All experiments are executed on a single v100 Nvidia GPU. To compare
computational costs, Table~\ref{tab:costs} reports the average training time
over all languages for each of the models. The training time for nmt-transfer
is the highest, followed by aux-nmt, then come the leaner auxiliary tasks. The
inference time of all the models for the SLU tasks is highly similar due to the
similar architecture (except for nmt-transfer requiring fairSeq a-priori).

\begin{table}
    \centering
    \resizebox{.625\columnwidth}{!}{
    \begin{tabular}{l r}
    \toprule
    Model & Time (minutes) \\
    \midrule
    base & 3\\
    nmt-transfer & 5,145\\
    aux-mlm & 220\\
    aux-nmt & 464\\
    aux-ud  & 57\\
    \bottomrule
    \end{tabular}}
    \caption{Average minutes to train a model, averaged over all languages and
both embeddings. For nmt-transfer we include the training of the NMT model.} 
    \label{tab:costs}
\end{table}

\subsection{Case Study: Improving on de-st}
Our lowest-resource language variety de-st is not included in either
embeddings, and the performance on it is generally low. To mitigate this, we
investigate whether a small amount of raw data could improve the aux-mlm model.
We scraped 23,572 tweets and 6,583 comments from ask.fm manually identified by
a native speaker, and used these as auxiliary data in the aux-mlm model.
Although this data is difficult to obtain and contains a mix including standard
German and others, it resulted in an increase from 49.9 to 56.2 in slot F1
scores and from 68.0 to 68.7 for intents, compared to using the German data in
aux-mlm, thereby largely outperforming the baseline. This shows that even small
amounts of data are highly beneficial in aux training, confirming results
of~\newcite{muller2020being}.

%% file: relwork.tex
\label{sec:relmodels}
For related datasets, we refer to \S~\ref{sec:datasets}; in this section we
will discuss different approaches on how to tackle cross-lingual SLU.  Work on
cross-lingual SLU can broadly be divided into two approaches, whether it is
based mainly on parallel data or multilingual representations. The first stream
of research focuses on generating training data in the target language with
machine translation and mapping the slot labels through attention or an
external word aligner.  The translation-based approach can be further improved
by filtering the resulting training
data~\cite{gaspers-etal-2018-selecting,do-gaspers-2019-cross}, post-fixing the
annotation by humans~\cite{castellucci_multi-lingual_2019}, or by using a
soft-alignment based on attention, which alleviates error propagation and
outperforms annotation projection using external word
aligners~\cite{xu_end--end_2020}. 

The second stream of research uses multilingual representations.
\newcite{upadhyay2018almost} use bilingual word embeddings based
on~\newcite{smith2017offline} in a bidirectional Long Short-Term Memory model
for zero-shot SLU. Recent work focuses on finding better multilingual
representations.  \newcite{schuster-etal-2019-cross-lingual} use a multilingual
machine translation encoder as word representations.
\newcite{liu-etal-2019-zero} propose refining the alignment of bilingual word
representations. The best performing variants use contextualized BERT
variants~\cite{chen_bert_2019,xu_end--end_2020}, which we depart from.

We propose a third, orthogonal line of research: joint target-language
auxiliary task learning. We hypothesize that jointly training on target
language auxiliary tasks  helps to learn properties of the target language
while learning a related task simultaneously. We frame masked language
modeling, Universal Dependency parsing and machine translation as new auxiliary
tasks for SLU.

Some work on SLU showed that syntax in graph convolution networks is beneficial
for slots~\cite{qin_multi-domain_2020}. Contemporary work shows that
high-resource English data helps target language modeling in sequential
transfer setups~\cite{phang2020english}. We focus on non-English target data
for joint SLU in a single cross-lingual multi-task model instead.

%% file: conclusion.tex
We introduced \name{}, a multilingual dataset for spoken language understanding
with 13 languages from 6 language families, including an unstudied German
dialect. \name{} includes a wide variety of intent types and homogenized
annotations. We propose non-English multi-task setups for zero-shot transfer to
learn the target language: masked language modeling, neural machine translation
and UD parsing. We compared the effect of these auxiliary tasks in two
settings.  Our results showed that masked language modeling led to the most
stable performance improvements; however, when a language is not seen during
pre-training, UD parsing led to an even larger performance increase. On the
intents, generating target language training data using machine translation was
outperforming all our proposed models, at a much higher computational cost
however. Our analysis further shows that  nmt-transfer struggles with span
detection. Given training time and availability trade-off, MLM multi-tasking is
a viable approach for SLU.

%% file: appendix.tex
\section*{Appendix}
\begin{table}[h!]\centering
  \begin{tabular}{l r }
    \toprule
    Parameter & Value  \\
    \midrule
    Optimizer                   & Adam \\
    $\beta_1$, $\beta_2$        & 0.9, 0.99 \\
    Dropout                     & 0.3 \\
    Epochs                      & 20 \\
    Batch size                  & 32 \\
    Learning rate               & 0.001\\
    Weight decay                & 0.01 \\
    LR scheduler      & slanted triangular \\
    Decay factor                & 0.38 \\
    Cut frac                    & 0.2  \\
\bottomrule
  \end{tabular}
  \caption{Hyperparameter setting used in the experiments}
  \label{tab:params}
\end{table}

\section{Hyperparameter settings} \label{app:hyperparams}
Hyperparameters follow the default setting of MaChAmp
0.2~\cite{vandergoot-etal-2020-machamp} with 20 epochs, as reported in
Table~\ref{tab:params}.

\section{Auxiliary Datasets} \label{app:aux-datasets}
Table~\ref{tab:auxSizes} reports the data sources for the treebanks and the
dataset sizes in number of words and sentences for both the treebanks and the
parallel data.

\begin{table*}
\centering
\begin{tabular}{l|lrr|rr}
\toprule
& \multicolumn{3}{c}{Universal Dependencies} & \multicolumn{2}{c}{Parallel data}\\
Lang. & Treebank & \#sents & \#words & \#sents & \#words \\
\midrule
ar & UD\_Arabic-PADT & 6,075 & 191,869 & 22,666,885 & 122,580,047 \\
da & UD\_Danish-DDT & 4,383 & 80,378 & 11,021,827 & 71,415,893 \\
de & UD\_German-GSD & 13,814 & 259,194 & 14,325,270 & 99,354,451 \\
en & UD\_English-EWT & 12,543 & 204,585 & 0 & 0 \\
es & UD\_Spanish-GSD & 14,187 & 375,149 & 61,434,251 & 415,369,072 \\
fr & UD\_French-GSD & 14,449 & 344,970 & 41,921,465 & 280,924,433 \\
hi & UD\_Hindi-HDTB & 13,304 & 281,057 & 93,016 & 620,929 \\
id & UD\_Indonesian-GSD & 4,477 & 97,531 & 5,370,460 & 30,758,822 \\
it & UD\_Italian-ISDT & 13,121 & 257,616 & 26,344,624 & 180,169,211 \\
ja & UD\_Japanese-GSD & 7,027 & 167,482 & 1,883,365 & 12,891,698 \\
kk & UD\_Kazakh-KTB & 1,047 & 9,872 & 595,060 & 10,058,764 \\
nl & UD\_Dutch-LassySmall & 5,787 & 75,080 & 28,835,007 & 196,968,670 \\
pt & UD\_Portuguese-GSD & 9,664 & 238,714 & 33,375,963 & 218,626,646 \\
sr & UD\_Serbian-SET & 3,328 & 74,259 & 22,319,620 & 133,297,245 \\
th & UD\_Thai-PUD & 1,000 & 22,322 & 3,281,533 & 4,332,396 \\
tr & UD\_Turkish-IMST & 3,664 & 36,822 & 45,788,547 & 229,132,015 \\
zh & UD\_Chinese-GSD & 3,997 & 98,616 & 9,475,118 & 89,458,907 \\
\midrule
\end{tabular}
\caption{Dataset sizes for auxiliary tasks, for our auxiliary setting we constrained the model to only use 10,000 sentences of the treebanks and 100,000 of the parallel data}
\label{tab:auxSizes}
\end{table*}

\section{Scores on auxiliary tasks} \label{app:aux-tasks}
Even though it was not our goal to improve the auxiliary tasks, performance on
these can still be relevant to analyze whether there is any correlation to
performance on the \name{} tasks. In Table~\ref{tab:auxScores}, we report the
full results for all tasks. These are the scores the correlations of
Figure~\ref{fig:correlations} are based on.

\begin{table*}
\centering
    \begin{tabular}{l| r r r r | r r r r }
    \toprule
        Lang & ud(avg.) & mlm & aux-nmt & nmt-transfer & ud(avg.) & mlm & aux-nmt & nmt-transfer \\ 
        \midrule
& mBERT & & & &  XLM15\\
        \midrule
ar & 88.12 & 1.81 & 11.97 & 16.78 & 88.51 & 3.21 & 1.01 & 16.78\\
da & 94.10 & 3.17 & 13.86 & 56.24 & 89.35 & 3.00 & 9.80 & 56.24\\
de & 93.00 & 1.87 & 14.22 & 25.93 & 92.83 & 2.63 & 1.19 & 25.93\\
en & 95.24 & 4.17 & --- & --- & 94.90 & 5.18 & --- & ---\\
id & 91.31 & 2.32 & 20.89 & 27.96 & 88.02 & 1.13 & 4.02 & 27.96\\
it & 96.83 & 4.03 & 8.98 & 44.73  & 95.46 & 2.52 & 7.17 & 44.73 \\
ja & 97.97 & 2.91 & 4.53 & 10.08 & 97.12 & 2.79 & 1.53 & 10.08 \\
kk & 70.55 & 1.48 & 2.67 & 0.00 & 52.08 & 2.65 & 0.09 & 0.00\\
nl & 93.84 & 3.81 & 11.66 & 53.43 & 90.14 & 2.59 & 9.50 & 53.43\\
sr & 94.18 & 3.54 & 9.78 & 35.50 & 89.81 & 4.14 & 7.25 & 35.50 \\
tr & 83.04 & 1.71 & 9.31 & 14.45 & 79.41 & 2.87 & 0.87 & 14.45\\
zh & 94.80 & 1.52 & 11.52 & 20.31  & 93.00 & 2.01 & 0.12 & 20.31\\
\bottomrule
    \end{tabular}
    \caption{Results on auxiliary tasks: for UD, we use the average over UPOS
accuracy, lemma accuracy, morphological feature accuracy (all features as 1
label) and depdendency LAS, for masked language modeling (mlm) we use
perplexity, and the last two columns (nmt) are bleu scores.}
    \label{tab:auxScores}
\end{table*}

\section{Standard Deviations}
\label{sec:stdev}
Standard deviations of our main results (Table~\ref{tab:mainResults}) are shown
in Table~\ref{tab:stdev}.

\begin{table*}[ht!]
\centering
\small
\setlength{\tabcolsep}{5pt}
\begin{tabular}{l| r | r r r r r r r r r r r r|r}
\toprule
mBERT & \textbf{en} & de-st & \textbf{de} & \textbf{da} & \textbf{nl} & \textbf{it} & \textbf{sr} & \textbf{id} & \textbf{ar} & \textbf{zh} & \textbf{kk} & \textbf{tr} & \textbf{ja}$^*$ & Avg.\\
lang2vec & --- & --- & 0.18 & 0.18 & 0.19 & 0.22 & 0.23 & 0.24 & 0.30 & 0.33 & 0.37 & 0.38 & 0.41 \\
\midrule
Slots\\
\midrule
base & 0.2 & 4.8 & 3.1 & 1.7 & 1.0 & 1.0 & 2.4 & 2.2 & 1.3 & 2.3 & 4.1 & 1.4 & 6.5 & 2.6 \\
nmt-transfer & 0.0 & 2.0 & 0.7 & 1.3 & 1.7 & 1.4 & 1.6 & 1.4 & 1.2 & 1.1 & 0.3 & 1.7 & 2.8 & 1.4 \\
aux-mlm & 0.4 & 1.8 & 0.9 & 1.9 & 1.3 & 0.9 & 2.0 & 2.3 & 3.0 & 4.0 & 2.0 & 3.1 & 3.1 & 2.2 \\
aux-nmt & 0.0 & 1.9 & 1.3 & 1.3 & 1.1 & 1.3 & 3.2 & 1.9 & 2.1 & 2.3 & 1.9 & 2.6 & 4.9 & 2.1 \\
aux-ud & 0.2 & 1.4 & 1.0 & 0.7 & 1.7 & 4.0 & 1.1 & 2.9 & 0.3 & 1.8 & 1.4 & 2.8 & 3.5 & 1.9 \\
\midrule
Intents\\
\midrule
base & 0.0 & 4.5 & 3.5 & 2.8 & 2.0 & 2.0 & 2.5 & 3.4 & 2.4 & 1.8 & 3.1 & 1.1 & 3.8 & 2.8 \\
nmt-transfer & 0.0 & 2.9 & 0.6 & 1.0 & 0.5 & 1.1 & 2.4 & 0.6 & 0.5 & 0.8 & 6.6 & 0.4 & 4.2 & 1.8 \\
aux-mlm & 0.2 & 2.6 & 2.6 & 2.0 & 2.6 & 4.2 & 4.6 & 7.2 & 3.4 & 2.3 & 4.5 & 5.9 & 8.9 & 4.2 \\
aux-nmt & 0.0 & 4.2 & 2.7 & 2.0 & 3.0 & 4.2 & 2.7 & 1.6 & 2.5 & 3.6 & 3.3 & 3.6 & 5.7 & 3.3 \\
aux-ud & 0.2 & 3.2 & 3.2 & 4.0 & 1.8 & 3.9 & 4.4 & 3.8 & 5.2 & 3.5 & 1.2 & 3.5 & 4.1 & 3.5 \\
\midrule[1pt]
XLM15 & \textbf{en} & de-st & \textbf{de} & da & nl & it & sr & id & \textbf{ar} & \textbf{zh} & kk & \textbf{tr} & ja$^*$ \\
\midrule
Slots\\
\midrule
base & 0.3 & 2.3 & 1.5 & 1.7 & 1.7 & 2.0 & 3.7 & 0.9 & 1.2 & 4.0 & 1.1 & 2.9 & 3.4 & 2.2 \\
nmt-transfer & 0.0 & 4.3 & 1.0 & 2.0 & 0.9 & 1.0 & 1.6 & 1.8 & 1.1 & 2.7 & 0.0 & 1.5 & 19.5 & 3.1 \\
aux-mlm & 0.4 & 3.1 & 0.4 & 4.0 & 3.6 & 1.4 & 1.2 & 3.1 & 1.4 & 2.7 & 1.3 & 2.0 & 8.5 & 2.7 \\
aux-nmt & 0.0 & 4.1 & 4.1 & 2.2 & 2.0 & 1.6 & 2.4 & 0.8 & 3.0 & 17.5 & 3.9 & 1.9 & 5.2 & 4.1 \\
aux-ud & 0.3 & 2.4 & 0.5 & 3.5 & 3.7 & 1.8 & 3.9 & 4.1 & 1.6 & 1.8 & 3.0 & 2.2 & 4.9 & 2.8 \\
\midrule
Intents\\
\midrule
base & 0.2 & 5.3 & 2.0 & 6.1 & 6.0 & 6.7 & 4.5 & 2.8 & 3.1 & 5.4 & 10.0 & 3.2 & 7.0 & 5.2 \\
nmt-transfer & 0.0 & 3.9 & 1.2 & 0.7 & 1.0 & 0.8 & 2.4 & 1.0 & 0.4 & 1.0 & 6.4 & 0.5 & 8.3 & 2.3 \\
aux-mlm & 0.1 & 2.1 & 2.0 & 4.6 & 6.0 & 2.6 & 7.9 & 8.3 & 4.2 & 1.3 & 6.0 & 7.0 & 4.6 & 4.7 \\
aux-nmt & 0.0 & 5.1 & 5.1 & 2.0 & 1.6 & 3.1 & 4.8 & 3.0 & 5.3 & 13.9 & 10.1 & 3.7 & 11.5 & 5.8 \\
aux-ud & 0.3 & 1.6 & 2.7 & 3.9 & 4.0 & 4.6 & 4.4 & 5.0 & 4.4 & 4.1 & 4.6 & 5.8 & 3.2 & 4.0 \\
\bottomrule
\end{tabular}
\caption{Standard deviation matching all results from our main results (Table~\ref{tab:mainResults})} 
\label{tab:stdev}
\end{table*}

\section{\name{} Data Statement} \label{app:datastatement}
Following~\cite{bender-friedman-2018-data}, the following outlines the data
statement for \name{}:

\textsc{A. CURATION RATIONALE} 
Collection of utterances intended to be used for digital assistants, generated
by crowd-workers. We selected a random sample from two much larger
sets~\cite{coucke_snips_2018,schuster-etal-2019-cross-lingual} which we
translated and annotated for slots and intents for the cross-lingual study of
SLU.

\textsc{B. LANGUAGE VARIETY} 
The English data was created by native English speakers and all translations
are translated by native speakers. We translated to the following languages
according to the iso 639-3 codes: 'deu', 'jpn', 'tur', 'nld', 'ita', 'dan',
'arb', 'kaz', 'srp','eng', 'ind', 'cmn'. South-tyrolean does not have an iso
693-3 language code.

\textsc{C. SPEAKER DEMOGRAPHIC} 
The original data is generated by crowd-workers and their demographics are
unknown.

\textsc{D. ANNOTATOR DEMOGRAPHIC} 
Translators and annotators are the same people. Their age ranges from 20 to 57,
with the majority being below 30, almost all annotators have a background in
NLP (except for Chinese, and one inter-annotator for Dutch). Most annotators
are currently doing a PhD, whereas there is one postdoc and two faculty.

\textsc{E. SPEECH SITUATION} 
The original data is generated in June 2017~\cite{coucke_snips_2018} and
probably in 2019~\cite{schuster-etal-2019-cross-lingual}. The crowd workers
were tasked to type sentences as how they would ask them in spoken form to a
digitial assistant given a topic (intent).

\textsc{F. TEXT CHARACTERISTICS} 
The genre of the data is determined by the set of supported intents:
\begin{verbatim}
AddToPlaylist
BookRestaurant
PlayMusic
RateBook
SearchCreativeWork
SearchScreeningEvent
alarm/cancel_alarm
alarm/modify_alarm
alarm/set_alarm
alarm/show_alarms
alarm/snooze_alarm
alarm/time_left_on_alarm
reminder/cancel_reminder
reminder/set_reminder
reminder/show_reminders
weather/find
\end{verbatim}

\textsc{I. PROVENANCE APPENDIX} 
The original datasets have been released with the following licenses:
\begin{itemize}
\item ~\newcite{schuster-etal-2019-cross-lingual}: CC-BY-SA license.
\item ~\newcite{coucke_snips_2018}: CC0 1.0 Universal
\end{itemize}
We use the CC-BY-SA license for our re-distribution of the data.
\\
\\

\section{Translation Guidelines} \label{app:trans-guidelines}
We aim to provide a fluent translation which is as similar (in meaning) as
possible to the original. In some cases translations naturally have more
distance, i.e. `7 pm' might translate to `7 in the evening' for languages in
which there is no equivalent for `pm'. The goal is to obtain sentences as they
could possibly be used in the target language. Some general guidelines:

\begin{itemize} 
\item In general, named entities are not translated, with the exception of
place names, like cities and countries. So names of playlists, persons etc.
stay the same, and things mentioned between quotes as well. In languages where
names are often transcribed differently (i.e. Serbian), this is done during
annotation.
\item In case of grammatical mistakes, they are kept (if possible) in the
target translation. 
\item We keep capitalization and punctuation as in the original data (if they
exist in the target language). 
\item Abbreviations not common in fluent discourse are expanded (e.g., Wed
$\mapsto$ "mercoledì", meds $\mapsto$ medicin), also words that do not exist in
the target language are paraphrased: `whats the high tomorrow' $\mapsto$ `whats
the maximum temperature tomorrow'.
\item Some things can not be translated directly. For example, the phrase `play
me X' does not exist in many languages. E.g. `me' might not be translated.
\item Possessive determiners (e.g. ``my'') should be preserved and translated
whenever possible.
\item For languages in which words are not separated by whitespace (i.e.
Japanese and Chinese), we ask the translator to include whitespaces at word
boundaries to simplify the annotation of the slots.
\end{itemize}

\clearpage
\section{Annotation Guidelines} \label{app:ann-guidelines}

\begin{figure}
\includegraphics[width=\columnwidth]{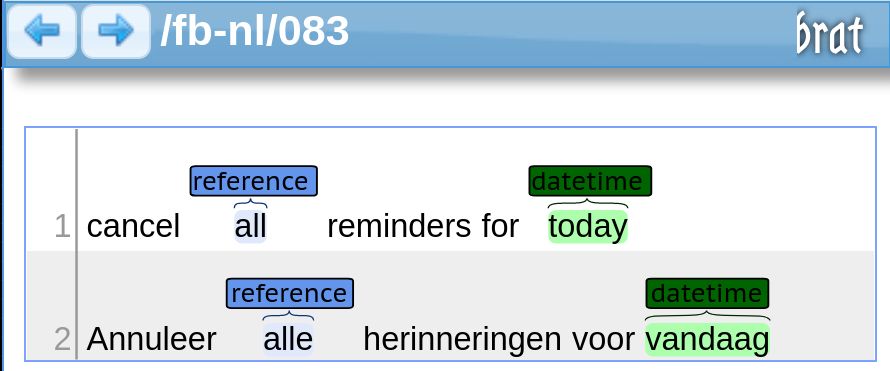}
\caption{Example showing the annotation of a Dutch instance in Brat}
\label{fig:brat}
\end{figure}

\begin{figure*}
\centering
\begin{tabular}{| l | l |}
\hline
1. & $\mbox{set alarm for } [\mbox{7 am}]_{datetime}$ \\

2. & $\mbox{set reminder from } [\mbox{5 to 6 am}]_{datetime}$ \\

3. & $\mbox{turn off } [\mbox{all}]_{reference} \mbox{ alarms for } [\mbox{tomorrow}]_{datetime}$ \\

4. & $\mbox{set a } [\mbox{daily}]_{recurring\_datetime} \mbox{ reminder}$ \\

5. & $\mbox{remind me to } [\mbox{submit my plan}]_{reminder\_todo} $ \\
 
6. & $\mbox{show } [\mbox{all my}]_{reference } \mbox{ alarms} $ \\

7. & $\mbox{will it be } [\mbox{sunny}]_{weather-attribute }  [\mbox{tomorrow}]_{datetime}$ \\

8. & $\mbox{i need an alarm for } [\mbox{5pm}]_{datetime} \mbox{ to remind me to } [\mbox{take my medicine}]_{reminder\_todo} $ \\
9 & $\mbox{schedule a } [\mbox{daily}]_{recurring\_datetime} \mbox{ alarm for } [\mbox{7:30pm}]_{recurring\_datetime} \mbox{ until deleted} $ \\
\hline
\end{tabular}
\caption{Example annotation for the slots. Slot-spans are indicated by square
brackets and their label is shown directly behind the span.}
\label{fig:examples2}
\end{figure*}

This section describes our annotation guidelines.  The aim of these guidelines
was to make annotations homogeneous across earlier efforts for which guidelines
were not available. Two major changes compared to the original annotations
include: i) to generalize labels to make them more broadly applicable to other
intent subtypes (an example is the recurrent datetime event from the Facebook
data~\cite{schuster-etal-2019-cross-lingual}, which was originally only applied
to reminders and not to alarms, as discussed below); ii) we drop annotations of
nouns as slots which are directly inferrable from the intent label (e.g. the
`reminder/noun' label was only applicable to nouns, but it was sometimes
expressed as a verb and hence annotations were missing; as they are already
annotated in the sentence-level intent slots, we drop such obvious slots). 

For the annotation, we use Brat~\cite{stenetorp-etal-2012-brat}, and provided
the annotators with the gold English annotation (see Figure~\ref{fig:brat}).
English annotation was conducted by three annotators who discussed and resolved
any initial disagreements. For annotation of the other languages, annotators
were instructed to follow the English annotation when possible to maintain
consistency.

Because no annotation guidelines were released with the original data, we
provide guidelines for our re-annotation of the slots below. Examples are shown
in Figure~\ref{fig:examples2}.

\paragraph{Spans} 
We exclude function words in the beginning of an NP or VP, like `for', `from'
in the examples above. An exception is when it is in the middle of a span as
contiguous slots are preferred, like in example 2. This is different from
previous releases of the  data~\cite{schuster-etal-2019-cross-lingual}, where
datetime included `for', `at', `to' and `on'. We decided to drop them to make
the annotations more homogeneous across slot labels, while capturing the core
(`head spans') of the slots. 

When two words of the same type occur sequentially, we annotate them as one
span. This happens both for datetime (example 2, [5 to 6 am]) as well as
reference (example 6, [all my]). Furthermore, we keep the annotation on the
word-level to simplify processing. If only a part of a word belongs to a label,
we annotate the whole word with that label.

\begin{figure*}
\centering
\begin{tabular}{l l l l l l l l l l l}
English: \\
\multicolumn{8}{l}{
Remind me to \colorbox{object}{call mom} \colorbox{movie}{today at 2 p.m.}} \vspace{.25cm} \\ 
German:\\
Erinnere & mich & \colorbox{object}{Mama} & \colorbox{movie}{heute} &  \colorbox{movie}{Nachmittag} & \colorbox{movie}{um} & \colorbox{movie}{2} & \colorbox{movie}{Uhr} & \colorbox{object}{anzurufen} \\
remind & me & mama & today & afternoon & at & 2 & o-clock & to call \\
\end{tabular}
\caption{Example of sentence-final verb in German. Green: reminder\_todo, Orange: datetime}
\label{fig:sentenceFinal}
\end{figure*}

\paragraph{Slot labels}
After our adaptations of the original labels, we annotate the following labels:
\begin{itemize}
    \item \texttt{datetime}: Indicating a date or a time. Only concrete times
are annotated (not, `until deleted', `what time' or `when'), and times relative
to other events are included (e.g. `after work', `later'). Non-concrete times,
like `until deleted' (example 9) are excluded.    
    \item \texttt{recurring\_datetime}: a recurring event, can be used for
alarms and reminders. This category prioritizes over datetime. Example: `make
alarm for [weekdays at 7 am]', if at least one recurring\_datetime exists in an
instance, all datetimes should be annotated as recurring\_datetime (even if
they are in different spans, see example 9).    
    \item \texttt{location}: describes a location; can be a proper noun (like
`New York') or a nominal or adjective referring to a location (`my area` , 'out
(outside)'). If a location is part of a reminder item, it is annotated as
\texttt{reminder/todo} instead.
    \item \texttt{reference:} modifies the scope of an alarm or reminder,
usually `my' or `all' used in front of the word `reminder(s)' or `alarm(s)'.
Multiple sequential references are annotated as one span (`cancel [all my]
alarms').
    \item \texttt{reminder/todo:} the item that should be reminded, the word
`to' should be excluded. In special cases, we also apply this for alarms (see
example 8).
    \item \texttt{weather/attribute:} A property that describes an aspect of
the weather; e.g. `cold', `rain', `temperature', `severe'. Also includes
weather-related items like `coat' and `umbrella' if used in relation to the
weather.
\end{itemize}

\subsection{Language-specific exceptions}

\paragraph{German/Dutch sentence-final verbs and split phrases} German and
Dutch allow for sentence-final verbs. For example, reminder todo slots might
end up being split in the German translation, as another phrase or slot might
be in between. See for an example Figure~\ref{fig:sentenceFinal}.  We annotate
the resulting non-continuous span for `call mum' as two separate
\texttt{reminder\_todo} entities. 

\paragraph{Handling of difficulties in Arabic compound morphology}

We encountered three special morphological cases during translating and
annotating Arabic samples, in which a word can be a compound of two segments:
\begin{enumerate}
  \setcode{utf8}

    \item Verbs can be prefixed by a preposition (e.g. to, for, by):\\
\<          ذكرني بشراء الحليب  >\\
           Remind me to buy milk \\
     In this case, we decided to annotate the whole compound word \<بشراء > (EN: to buy milk) as a reminder. 
    \item Nouns can be suffixed with a possessive determiner (e.g. my, our): \\
\<           اغلق منبهي >\\
           Turn off my alarm \\
     In this case we decided to annotated the whole compound word \<منبهي> (EN: ``my alarm'') as reference-part. This label was only used for Arabic, and is converted to reference during evaluation.
    \item Singular and dual nouns can be written as one word without the need to modify it with the numbers one or two: \\
           two stars $\mapsto$ \<  نجمتين >\\
           one star $\mapsto$ \< نجمة >\\
\end{enumerate}

\paragraph{Serbian orthographic transcription of foreign named entities} 
The rules of standard Serbian spelling dictate that named entities from other
languages be transcribed orthographically into Serbian. Since Serbian
orthography is near-perfectly phonemic, this means that the transcription of
named entities will be highly dependent on the source pronunciation (although
not solely determined by it). As a result, different transcription rules apply
to not only different types of named entities but also different source
languages.

When translating to Serbian, we recognize three groups of named entities that
require different rules for translation and annotation: 

\begin{enumerate}
    \item \textit{Common named entities, such as well-known place names, names of
historical and public figures, as well as names of popular literary, visual,
and musical works.} Such named entities are simply translated into their
established Serbian equivalents.
    \item \textit{Less common named entities, such as lesser-known place names or
names of public figures.} When translating these entities, we follow the
pronunciation and transcription rules specific to the source language, which
means that we have to look up the origin and native pronunciation of all
unfamiliar entities. However, since many of these are unlikely to be found in
Serbian texts or corpora, we keep the source transcription (in English) in
square brackets for future reference. We annotate the source and its Serbian
translation as two separate spans with the same label.
    \item \textit{Names of songs, playlists, video games, and lesser-known films
and TV programs.} These were neither translated nor transcribed, but left as
they are in the source text, as such named entities are not commonly
transcribed and have no official translation in Serbian.
\end{enumerate}

%% file: main.bbl
\begin{thebibliography}{52}
\expandafter\ifx\csname natexlab\endcsname\relax\def\natexlab#1{#1}\fi

\bibitem[{Bahdanau et~al.(2015)Bahdanau, Cho, and Bengio}]{bahdanau2014neural}
Dzmitry Bahdanau, Kyunghyun Cho, and Yoshua Bengio. 2015.
\newblock Neural machine translation by jointly learning to align and
  translate.
\newblock \emph{Proceedings of International Conference on Learning
  Representations (ICLR)}.

\bibitem[{Barrault et~al.(2019)Barrault, Bojar, Costa-juss{\`a}, Federmann,
  Fishel, Graham, Haddow, Huck, Koehn, Malmasi, Monz, M{\"u}ller, Pal, Post,
  and Zampieri}]{barrault-etal-2019-findings}
Lo{\"\i}c Barrault, Ond{\v{r}}ej Bojar, Marta~R. Costa-juss{\`a}, Christian
  Federmann, Mark Fishel, Yvette Graham, Barry Haddow, Matthias Huck, Philipp
  Koehn, Shervin Malmasi, Christof Monz, Mathias M{\"u}ller, Santanu Pal, Matt
  Post, and Marcos Zampieri. 2019.
\newblock \href {https://doi.org/10.18653/v1/W19-5301} {Findings of the 2019
  conference on machine translation ({WMT}19)}.
\newblock In \emph{Proceedings of the Fourth Conference on Machine Translation
  (Volume 2: Shared Task Papers, Day 1)}, pages 1--61, Florence, Italy.
  Association for Computational Linguistics.

\bibitem[{Bellomaria et~al.(2019)Bellomaria, Castellucci, Favalli, and
  Romagnoli}]{bellomaria2019almawave}
Valentina Bellomaria, Giuseppe Castellucci, Andrea Favalli, and Raniero
  Romagnoli. 2019.
\newblock Almawave-slu: a new dataset for {SLU} in {I}talian.
\newblock In \emph{Proceedings of the Sixth Italian Conference on Computational
  Linguistics}.

\bibitem[{Bender and Friedman(2018)}]{bender-friedman-2018-data}
Emily~M. Bender and Batya Friedman. 2018.
\newblock \href {https://doi.org/10.1162/tacl_a_00041} {Data statements for
  natural language processing: Toward mitigating system bias and enabling
  better science}.
\newblock \emph{Transactions of the Association for Computational Linguistics},
  6:587--604.

\bibitem[{Bonferroni(1936)}]{bonferroni1936teoria}
Carlo Bonferroni. 1936.
\newblock Teoria statistica delle classi e calcolo delle probabilita.
\newblock \emph{Pubblicazioni del R Istituto Superiore di Scienze Economiche e
  Commericiali di Firenze}, 8:3--62.

\bibitem[{Castellucci et~al.(2019)Castellucci, Bellomaria, Favalli, and
  Romagnoli}]{castellucci_multi-lingual_2019}
Giuseppe Castellucci, Valentina Bellomaria, Andrea Favalli, and Raniero
  Romagnoli. 2019.
\newblock Multi-lingual intent detection and slot filling in a joint
  {BERT}-based model.
\newblock \emph{arXiv:1907.02884 [cs]}.
\newblock ArXiv: 1907.02884.

\bibitem[{Cettolo et~al.(2016)Cettolo, Jan, Sebastian, Bentivogli, Cattoni, and
  Federico}]{cettolo2016iwslt}
Mauro Cettolo, Niehues Jan, St{\"u}ker Sebastian, Luisa Bentivogli, Roldano
  Cattoni, and Marcello Federico. 2016.
\newblock \href
  {https://workshop2016.iwslt.org/downloads/IWSLT_2016_evaluation_overview.pdf}
  {The {IWSLT} 2016 evaluation campaign}.
\newblock In \emph{International Workshop on Spoken Language Translation}.

\bibitem[{Chen et~al.(2019)Chen, Zhuo, and Wang}]{chen_bert_2019}
Qian Chen, Zhu Zhuo, and Wen Wang. 2019.
\newblock \href {https://arxiv.org/pdf/1902.10909.pdf} {Bert for joint intent
  classification and slot filling}.
\newblock \emph{arXiv preprint: 1902.10909}.

\bibitem[{Cho et~al.(2014)Cho, van Merri{\"e}nboer, Gulcehre, Bahdanau,
  Bougares, Schwenk, and Bengio}]{cho-etal-2014-learning}
Kyunghyun Cho, Bart van Merri{\"e}nboer, Caglar Gulcehre, Dzmitry Bahdanau,
  Fethi Bougares, Holger Schwenk, and Yoshua Bengio. 2014.
\newblock \href {https://doi.org/10.3115/v1/D14-1179} {Learning phrase
  representations using {RNN} encoder{--}decoder for statistical machine
  translation}.
\newblock In \emph{Proceedings of the 2014 Conference on Empirical Methods in
  Natural Language Processing ({EMNLP})}, pages 1724--1734, Doha, Qatar.
  Association for Computational Linguistics.

\bibitem[{Clinchant et~al.(2019)Clinchant, Jung, and
  Nikoulina}]{clinchant-etal-2019-use}
Stephane Clinchant, Kweon~Woo Jung, and Vassilina Nikoulina. 2019.
\newblock \href {https://doi.org/10.18653/v1/D19-5611} {On the use of {BERT}
  for neural machine translation}.
\newblock In \emph{Proceedings of the 3rd Workshop on Neural Generation and
  Translation}, pages 108--117, Hong Kong. Association for Computational
  Linguistics.

\bibitem[{Conneau et~al.(2020)Conneau, Khandelwal, Goyal, Chaudhary, Wenzek,
  Guzm{\'a}n, Grave, Ott, Zettlemoyer, and
  Stoyanov}]{conneau-etal-2020-unsupervised}
Alexis Conneau, Kartikay Khandelwal, Naman Goyal, Vishrav Chaudhary, Guillaume
  Wenzek, Francisco Guzm{\'a}n, Edouard Grave, Myle Ott, Luke Zettlemoyer, and
  Veselin Stoyanov. 2020.
\newblock \href {https://doi.org/10.18653/v1/2020.acl-main.747} {Unsupervised
  cross-lingual representation learning at scale}.
\newblock In \emph{Proceedings of the 58th Annual Meeting of the Association
  for Computational Linguistics}, pages 8440--8451, Online. Association for
  Computational Linguistics.

\bibitem[{Conneau and Lample(2019)}]{conneau2019cross}
Alexis Conneau and Guillaume Lample. 2019.
\newblock \href
  {https://proceedings.neurips.cc/paper/2019/file/c04c19c2c2474dbf5f7ac4372c5b9af1-Paper.pdf}
  {Cross-lingual language model pretraining}.
\newblock In \emph{Advances in Neural Information Processing Systems}, pages
  7059--7069.

\bibitem[{Coucke et~al.(2018)Coucke, Saade, Ball, Bluche, Caulier, Leroy,
  Doumouro, Gisselbrecht, Caltagirone, Lavril et~al.}]{coucke_snips_2018}
Alice Coucke, Alaa Saade, Adrien Ball, Th{\'e}odore Bluche, Alexandre Caulier,
  David Leroy, Cl{\'e}ment Doumouro, Thibault Gisselbrecht, Francesco
  Caltagirone, Thibaut Lavril, et~al. 2018.
\newblock \href {http://arxiv.org/abs/1805.10190} {Snips voice platform: an
  embedded spoken language understanding system for private-by-design voice
  interfaces}.
\newblock \emph{arXiv preprint: 1805.10190}.

\bibitem[{Devlin et~al.(2019)Devlin, Chang, Lee, and
  Toutanova}]{devlin-etal-2019-bert}
Jacob Devlin, Ming-Wei Chang, Kenton Lee, and Kristina Toutanova. 2019.
\newblock \href {https://doi.org/10.18653/v1/N19-1423} {{BERT}: Pre-training of
  deep bidirectional transformers for language understanding}.
\newblock In \emph{Proceedings of the 2019 Conference of the North {A}merican
  Chapter of the Association for Computational Linguistics: Human Language
  Technologies, Volume 1 (Long and Short Papers)}, pages 4171--4186,
  Minneapolis, Minnesota. Association for Computational Linguistics.

\bibitem[{Do and Gaspers(2019)}]{do-gaspers-2019-cross}
Quynh Do and Judith Gaspers. 2019.
\newblock \href {https://doi.org/10.18653/v1/D19-1153} {Cross-lingual transfer
  learning with data selection for large-scale spoken language understanding}.
\newblock In \emph{Proceedings of the 2019 Conference on Empirical Methods in
  Natural Language Processing and the 9th International Joint Conference on
  Natural Language Processing (EMNLP-IJCNLP)}, pages 1455--1460, Hong Kong,
  China. Association for Computational Linguistics.

\bibitem[{Dror et~al.(2019)Dror, Shlomov, and Reichart}]{dror2019deep}
Rotem Dror, Segev Shlomov, and Roi Reichart. 2019.
\newblock \href {https://doi.org/10.18653/v1/p19-1266} {Deep dominance - how to
  properly compare deep neural models}.
\newblock In \emph{Proceedings of the 57th Conference of the Association for
  Computational Linguistics, {ACL} 2019, Florence, Italy, July 28- August 2,
  2019, Volume 1: Long Papers}, pages 2773--2785. Association for Computational
  Linguistics.

\bibitem[{Einolghozati et~al.(2021)Einolghozati, Arora, Sainz-Maza~Lecanda,
  Kumar, and Gupta}]{einolghozati2021volumen}
Arash Einolghozati, Abhinav Arora, Lorena Sainz-Maza~Lecanda, Anuj Kumar, and
  Sonal Gupta. 2021.
\newblock \href {https://www.aclweb.org/anthology/2021.eacl-main.87} {El
  volumen louder por favor: Code-switching in task-oriented semantic parsing}.
\newblock In \emph{Proceedings of the 16th Conference of the European Chapter
  of the Association for Computational Linguistics: Main Volume}, pages
  1009--1021, Online. Association for Computational Linguistics.

\bibitem[{Fleiss(1971)}]{fleiss1971measuring}
Joseph~L Fleiss. 1971.
\newblock Measuring nominal scale agreement among many raters.
\newblock \emph{Psychological bulletin}, 76(5):378.

\bibitem[{Gardner et~al.(2018)Gardner, Grus, Neumann, Tafjord, Dasigi, Liu,
  Peters, Schmitz, and Zettlemoyer}]{Gardner2017AllenNLP}
Matt Gardner, Joel Grus, Mark Neumann, Oyvind Tafjord, Pradeep Dasigi,
  Nelson~F. Liu, Matthew Peters, Michael Schmitz, and Luke Zettlemoyer. 2018.
\newblock \href {https://doi.org/10.18653/v1/W18-2501} {{A}llen{NLP}: A deep
  semantic natural language processing platform}.
\newblock In \emph{Proceedings of Workshop for {NLP} Open Source Software
  ({NLP}-{OSS})}, pages 1--6, Melbourne, Australia. Association for
  Computational Linguistics.

\bibitem[{Gaspers et~al.(2018)Gaspers, Karanasou, and
  Chatterjee}]{gaspers-etal-2018-selecting}
Judith Gaspers, Penny Karanasou, and Rajen Chatterjee. 2018.
\newblock \href {https://doi.org/10.18653/v1/N18-3017} {Selecting
  machine-translated data for quick bootstrapping of a natural language
  understanding system}.
\newblock In \emph{Proceedings of the 2018 Conference of the North {A}merican
  Chapter of the Association for Computational Linguistics: Human Language
  Technologies, Volume 3 (Industry Papers)}, pages 137--144, New Orleans -
  Louisiana. Association for Computational Linguistics.

\bibitem[{Godwin et~al.(2016)Godwin, Stenetorp, and Riedel}]{godwin2016deep}
Jonathan Godwin, Pontus Stenetorp, and Sebastian Riedel. 2016.
\newblock \href {https://arxiv.org/pdf/1612.09113} {Deep semi-supervised
  learning with linguistically motivated sequence labeling task hierarchies}.
\newblock \emph{arXiv preprint: 1612.09113}.

\bibitem[{Goodfellow et~al.(2014)Goodfellow, Mirza, Xiao, Courville, and
  Bengio}]{goodfellow2013empirical}
Ian~J Goodfellow, Mehdi Mirza, Da~Xiao, Aaron Courville, and Yoshua Bengio.
  2014.
\newblock An empirical investigation of catastrophic forgetting in
  gradient-based neural networks.
\newblock \emph{Proceedings of International Conference on Learning
  Representations (ICLR)}.

\bibitem[{Gururangan et~al.(2020)Gururangan, Marasovi{\'c}, Swayamdipta, Lo,
  Beltagy, Downey, and Smith}]{gururangan-etal-2020-dont}
Suchin Gururangan, Ana Marasovi{\'c}, Swabha Swayamdipta, Kyle Lo, Iz~Beltagy,
  Doug Downey, and Noah~A. Smith. 2020.
\newblock \href {https://doi.org/10.18653/v1/2020.acl-main.740} {Don{'}t stop
  pretraining: Adapt language models to domains and tasks}.
\newblock In \emph{Proceedings of the 58th Annual Meeting of the Association
  for Computational Linguistics}, pages 8342--8360, Online. Association for
  Computational Linguistics.

\bibitem[{Hammarström and Nordhoff(2011)}]{glottlog}
Harald Hammarström and Sebastian Nordhoff. 2011.
\newblock \href
  {https://www.journals.uio.no/index.php/osla/article/view/75/199} {Langdoc:
  Bibliographic infrastructure for linguistic typology}.
\newblock \emph{Oslo Studies in Language}, 3(2):31--43.

\bibitem[{Hashimoto et~al.(2017)Hashimoto, Xiong, Tsuruoka, and
  Socher}]{hashimoto-etal-2017-joint}
Kazuma Hashimoto, Caiming Xiong, Yoshimasa Tsuruoka, and Richard Socher. 2017.
\newblock \href {https://doi.org/10.18653/v1/D17-1206} {A joint many-task
  model: Growing a neural network for multiple {NLP} tasks}.
\newblock In \emph{Proceedings of the 2017 Conference on Empirical Methods in
  Natural Language Processing}, pages 1923--1933, Copenhagen, Denmark.
  Association for Computational Linguistics.

\bibitem[{He et~al.(2013)He, Deng, Hakkani-Tur, and Tur}]{he2013multi}
Xiaodong He, Li~Deng, Dilek Hakkani-Tur, and Gokhan Tur. 2013.
\newblock \href
  {https://ieeexplore.ieee.org/stamp/stamp.jsp?tp=&arnumber=6639292}
  {Multi-style adaptive training for robust cross-lingual spoken language
  understanding}.
\newblock In \emph{2013 IEEE International Conference on Acoustics, Speech and
  Signal Processing}, pages 8342--8346. IEEE.

\bibitem[{Hemphill et~al.(1990)Hemphill, Godfrey, and
  Doddington}]{hemphill1990atis}
Charles~T. Hemphill, John~J. Godfrey, and George~R. Doddington. 1990.
\newblock \href {https://www.aclweb.org/anthology/H90-1021} {The {ATIS} spoken
  language systems pilot corpus}.
\newblock In \emph{Speech and Natural Language: Proceedings of a Workshop Held
  at Hidden Valley, {P}ennsylvania, June 24-27,1990}.

\bibitem[{Imamura and Sumita(2019)}]{imamura-sumita-2019-recycling}
Kenji Imamura and Eiichiro Sumita. 2019.
\newblock \href {https://doi.org/10.18653/v1/D19-5603} {Recycling a pre-trained
  {BERT} encoder for neural machine translation}.
\newblock In \emph{Proceedings of the 3rd Workshop on Neural Generation and
  Translation}, pages 23--31, Hong Kong. Association for Computational
  Linguistics.

\bibitem[{Keung et~al.(2020)Keung, Lu, Salazar, and
  Bhardwaj}]{keung-etal-2020-dont}
Phillip Keung, Yichao Lu, Julian Salazar, and Vikas Bhardwaj. 2020.
\newblock \href {https://www.aclweb.org/anthology/2020.emnlp-main.40} {Don{'}t
  use {E}nglish dev: On the zero-shot cross-lingual evaluation of contextual
  embeddings}.
\newblock In \emph{Proceedings of the 2020 Conference on Empirical Methods in
  Natural Language Processing (EMNLP)}, pages 549--554, Online. Association for
  Computational Linguistics.

\bibitem[{Lison and Tiedemann(2016)}]{lison-tiedemann-2016-opensubtitles2016}
Pierre Lison and J{\"o}rg Tiedemann. 2016.
\newblock \href {https://www.aclweb.org/anthology/L16-1147}
  {{O}pen{S}ubtitles2016: Extracting large parallel corpora from movie and {TV}
  subtitles}.
\newblock In \emph{Proceedings of the Tenth International Conference on
  Language Resources and Evaluation ({LREC}'16)}, pages 923--929,
  Portoro{\v{z}}, Slovenia. European Language Resources Association (ELRA).

\bibitem[{Littell et~al.(2017)Littell, Mortensen, Lin, Kairis, Turner, and
  Levin}]{littell-etal-2017-uriel}
Patrick Littell, David~R. Mortensen, Ke~Lin, Katherine Kairis, Carlisle Turner,
  and Lori Levin. 2017.
\newblock \href {https://www.aclweb.org/anthology/E17-2002} {{URIEL} and
  lang2vec: Representing languages as typological, geographical, and
  phylogenetic vectors}.
\newblock In \emph{Proceedings of the 15th Conference of the {E}uropean Chapter
  of the Association for Computational Linguistics: Volume 2, Short Papers},
  pages 8--14, Valencia, Spain. Association for Computational Linguistics.

\bibitem[{Liu et~al.(2019{\natexlab{a}})Liu, Ott, Goyal, Du, Joshi, Chen, Levy,
  Lewis, Zettlemoyer, and Stoyanov}]{liu2019roberta}
Yinhan Liu, Myle Ott, Naman Goyal, Jingfei Du, Mandar Joshi, Danqi Chen, Omer
  Levy, Mike Lewis, Luke Zettlemoyer, and Veselin Stoyanov. 2019{\natexlab{a}}.
\newblock \href {https://arxiv.org/pdf/1907.11692} {{RoBERTa}: A robustly
  optimized bert pretraining approach}.
\newblock \emph{arXiv preprint: 1907.11692}.

\bibitem[{Liu et~al.(2019{\natexlab{b}})Liu, Shin, Xu, Winata, Xu, Madotto, and
  Fung}]{liu-etal-2019-zero}
Zihan Liu, Jamin Shin, Yan Xu, Genta~Indra Winata, Peng Xu, Andrea Madotto, and
  Pascale Fung. 2019{\natexlab{b}}.
\newblock \href {https://doi.org/10.18653/v1/D19-1129} {Zero-shot cross-lingual
  dialogue systems with transferable latent variables}.
\newblock In \emph{Proceedings of the 2019 Conference on Empirical Methods in
  Natural Language Processing and the 9th International Joint Conference on
  Natural Language Processing (EMNLP-IJCNLP)}, pages 1297--1303, Hong Kong,
  China. Association for Computational Linguistics.

\bibitem[{McCloskey and Cohen(1989)}]{mccloskey1989catastrophic}
Michael McCloskey and Neal~J Cohen. 1989.
\newblock Catastrophic interference in connectionist networks: The sequential
  learning problem.
\newblock In \emph{Psychology of learning and motivation}, volume~24, pages
  109--165. Elsevier.

\bibitem[{Muller et~al.(2021)Muller, Anastasopoulos, Sagot, and
  Seddah}]{muller2020being}
Benjamin Muller, Antonis Anastasopoulos, Benoît Sagot, and Djamé Seddah.
  2021.
\newblock When being unseen from mbert is just the beginning: Handling new
  languages with multilingual language models.
\newblock In \emph{Proceedings of the 2021 Conference of the North {A}merican
  Chapter of the Association for Computational Linguistics: Human Language
  Technologies, Volume 1 (Long and Short Papers)}. Association for
  Computational Linguistics.

\bibitem[{Nivre et~al.(2020)Nivre, de~Marneffe, Ginter, Haji{\v{c}}, Manning,
  Pyysalo, Schuster, Tyers, and Zeman}]{nivre-etal-2020-universal}
Joakim Nivre, Marie-Catherine de~Marneffe, Filip Ginter, Jan Haji{\v{c}},
  Christopher~D. Manning, Sampo Pyysalo, Sebastian Schuster, Francis Tyers, and
  Daniel Zeman. 2020.
\newblock \href {https://www.aclweb.org/anthology/2020.lrec-1.497} {{U}niversal
  {D}ependencies v2: An evergrowing multilingual treebank collection}.
\newblock In \emph{Proceedings of the 12th Language Resources and Evaluation
  Conference}, pages 4034--4043, Marseille, France. European Language Resources
  Association.

\bibitem[{Ott et~al.(2019)Ott, Edunov, Baevski, Fan, Gross, Ng, Grangier, and
  Auli}]{ott2019fairseq}
Myle Ott, Sergey Edunov, Alexei Baevski, Angela Fan, Sam Gross, Nathan Ng,
  David Grangier, and Michael Auli. 2019.
\newblock \href {https://doi.org/10.18653/v1/N19-4009} {fairseq: A fast,
  extensible toolkit for sequence modeling}.
\newblock In \emph{Proceedings of the 2019 Conference of the North {A}merican
  Chapter of the Association for Computational Linguistics (Demonstrations)},
  pages 48--53, Minneapolis, Minnesota. Association for Computational
  Linguistics.

\bibitem[{Phang et~al.(2020)Phang, Calixto, Htut, Pruksachatkun, Liu, Vania,
  Kann, and Bowman}]{phang2020english}
Jason Phang, Iacer Calixto, Phu~Mon Htut, Yada Pruksachatkun, Haokun Liu, Clara
  Vania, Katharina Kann, and Samuel~R. Bowman. 2020.
\newblock \href {https://www.aclweb.org/anthology/2020.aacl-main.56} {{E}nglish
  intermediate-task training improves zero-shot cross-lingual transfer too}.
\newblock In \emph{Proceedings of the 1st Conference of the Asia-Pacific
  Chapter of the Association for Computational Linguistics and the 10th
  International Joint Conference on Natural Language Processing}, pages
  557--575, Suzhou, China. Association for Computational Linguistics.

\bibitem[{Qin et~al.(2020)Qin, Ni, Zhang, Che, Li, and
  Liu}]{qin_multi-domain_2020}
Libo Qin, Minheng Ni, Yue Zhang, Wanxiang Che, Yangming Li, and Ting Liu. 2020.
\newblock Multi-domain spoken language understanding using domain- and
  task-aware parameterization.
\newblock \emph{arXiv:2004.14871 [cs]}.
\newblock ArXiv: 2004.14871.

\bibitem[{Schuster et~al.(2019)Schuster, Gupta, Shah, and
  Lewis}]{schuster-etal-2019-cross-lingual}
Sebastian Schuster, Sonal Gupta, Rushin Shah, and Mike Lewis. 2019.
\newblock \href {https://doi.org/10.18653/v1/N19-1380} {Cross-lingual transfer
  learning for multilingual task oriented dialog}.
\newblock In \emph{Proceedings of the 2019 Conference of the North {A}merican
  Chapter of the Association for Computational Linguistics: Human Language
  Technologies, Volume 1 (Long and Short Papers)}, pages 3795--3805,
  Minneapolis, Minnesota. Association for Computational Linguistics.

\bibitem[{Sennrich et~al.(2016)Sennrich, Haddow, and
  Birch}]{sennrich-etal-2016-neural}
Rico Sennrich, Barry Haddow, and Alexandra Birch. 2016.
\newblock \href {https://doi.org/10.18653/v1/P16-1162} {Neural machine
  translation of rare words with subword units}.
\newblock In \emph{Proceedings of the 54th Annual Meeting of the Association
  for Computational Linguistics (Volume 1: Long Papers)}, pages 1715--1725,
  Berlin, Germany. Association for Computational Linguistics.

\bibitem[{Smith et~al.(2017)Smith, Turban, Hamblin, and
  Hammerla}]{smith2017offline}
Samuel~L Smith, David~HP Turban, Steven Hamblin, and Nils~Y Hammerla. 2017.
\newblock Offline bilingual word vectors, orthogonal transformations and the
  inverted softmax.
\newblock In \emph{Proceedings of International Conference on Learning
  Representations (ICLR)}.

\bibitem[{Sowa{\'n}ski and Janicki(2020)}]{sowanski2020leyzer}
Marcin Sowa{\'n}ski and Artur Janicki. 2020.
\newblock \href
  {https://link.springer.com/chapter/10.1007/978-3-030-58323-1_51} {Leyzer: A
  dataset for multilingual virtual assistants}.
\newblock In \emph{International Conference on Text, Speech, and Dialogue},
  pages 477--486. Springer.

\bibitem[{Stenetorp et~al.(2012)Stenetorp, Pyysalo, Topi{\'c}, Ohta, Ananiadou,
  and Tsujii}]{stenetorp-etal-2012-brat}
Pontus Stenetorp, Sampo Pyysalo, Goran Topi{\'c}, Tomoko Ohta, Sophia
  Ananiadou, and Jun{'}ichi Tsujii. 2012.
\newblock \href {https://www.aclweb.org/anthology/E12-2021} {brat: a web-based
  tool for {NLP}-assisted text annotation}.
\newblock In \emph{Proceedings of the Demonstrations at the 13th Conference of
  the {E}uropean Chapter of the Association for Computational Linguistics},
  pages 102--107, Avignon, France. Association for Computational Linguistics.

\bibitem[{Sutskever et~al.(2014)Sutskever, Vinyals, and
  Le}]{sutskever2014sequence}
Ilya Sutskever, Oriol Vinyals, and Quoc~V Le. 2014.
\newblock \href
  {https://papers.nips.cc/paper/2014/file/a14ac55a4f27472c5d894ec1c3c743d2-Paper.pdf}
  {Sequence to sequence learning with neural networks}.
\newblock In \emph{Advances in neural information processing systems}, pages
  3104--3112.

\bibitem[{Tiedemann(2012)}]{tiedemann-2012-parallel}
J{\"o}rg Tiedemann. 2012.
\newblock \href
  {http://www.lrec-conf.org/proceedings/lrec2012/pdf/463_Paper.pdf} {Parallel
  data, tools and interfaces in {OPUS}}.
\newblock In \emph{Proceedings of the Eighth International Conference on
  Language Resources and Evaluation ({LREC}'12)}, pages 2214--2218, Istanbul,
  Turkey. European Language Resources Association (ELRA).

\bibitem[{Ulmer(2021)}]{dennis_ulmer_2021_4638709}
Dennis Ulmer. 2021.
\newblock \href {https://doi.org/10.5281/zenodo.4638709} {deep-significance:
  Easy and better significance testing for deep neural networks}.
\newblock Https://github.com/Kaleidophon/deep-significance.

\bibitem[{Upadhyay et~al.(2018)Upadhyay, Faruqui, T{\"u}r, Dilek, and
  Heck}]{upadhyay2018almost}
Shyam Upadhyay, Manaal Faruqui, Gokhan T{\"u}r, Hakkani-T{\"u}r Dilek, and
  Larry Heck. 2018.
\newblock \href
  {https://ieeexplore.ieee.org/stamp/stamp.jsp?tp=&arnumber=8461905}
  {({A}lmost) zero-shot cross-lingual spoken language understanding}.
\newblock In \emph{2018 IEEE International Conference on Acoustics, Speech and
  Signal Processing (ICASSP)}, pages 6034--6038. IEEE.

\bibitem[{van~der Goot et~al.(2021)van~der Goot, {\"U}st{\"u}n, Ramponi,
  Sharaf, and Plank}]{vandergoot-etal-2020-machamp}
Rob van~der Goot, Ahmet {\"U}st{\"u}n, Alan Ramponi, Ibrahim Sharaf, and
  Barbara Plank. 2021.
\newblock \href {https://www.aclweb.org/anthology/2021.eacl-demos.22} {Massive
  choice, ample tasks ({M}a{C}h{A}mp): A toolkit for multi-task learning in
  {NLP}}.
\newblock In \emph{Proceedings of the 16th Conference of the European Chapter
  of the Association for Computational Linguistics: System Demonstrations},
  pages 176--197, Online. Association for Computational Linguistics.

\bibitem[{Vaswani et~al.(2017)Vaswani, Shazeer, Parmar, Uszkoreit, Jones,
  Gomez, Kaiser, and Polosukhin}]{vaswani2017attention}
Ashish Vaswani, Noam Shazeer, Niki Parmar, Jakob Uszkoreit, Llion Jones,
  Aidan~N Gomez, {\L}ukasz Kaiser, and Illia Polosukhin. 2017.
\newblock \href
  {https://papers.nips.cc/paper/2017/file/3f5ee243547dee91fbd053c1c4a845aa-Paper.pdf}
  {Attention is all you need}.
\newblock In \emph{Advances in neural information processing systems}, pages
  5998--6008.

\bibitem[{Xingkun~Liu and Rieser(2019)}]{XLiu.etal:IWSDS2019}
Pawel~Swietojanski Xingkun~Liu, Arash~Eshghi and Verena Rieser. 2019.
\newblock Benchmarking natural language understanding services for building
  conversational agents.
\newblock In \emph{Proceedings of the Tenth International Workshop on Spoken
  Dialogue Systems Technology (IWSDS)}, Ortigia, Siracusa (SR), Italy.
  Springer.

\bibitem[{Xu et~al.(2020)Xu, Haider, and Mansour}]{xu_end--end_2020}
Weijia Xu, Batool Haider, and Saab Mansour. 2020.
\newblock \href {https://doi.org/10.18653/v1/2020.emnlp-main.410} {End-to-end
  slot alignment and recognition for cross-lingual {NLU}}.
\newblock In \emph{Proceedings of the 2020 Conference on Empirical Methods in
  Natural Language Processing (EMNLP)}, pages 5052--5063, Online. Association
  for Computational Linguistics.

\end{thebibliography}
